%% file: main.tex
\documentclass[11pt]{article}
\usepackage[margin=1in]{geometry}
\input{math_commands.tex}

\usepackage[
            CJKbookmarks=true,
            bookmarksnumbered=true,
            bookmarksopen=true,
            colorlinks=true,
            citecolor=blue,
            linkcolor=blue,
            anchorcolor=red,
            urlcolor=blue
            ]{hyperref}
\usepackage{url}

\usepackage{amsmath,amsthm,amsfonts,amssymb}
\usepackage{tikz}
\usepackage{prettyref}
\usepackage{graphicx}
\usepackage{subcaption}
\usepackage{wrapfig}
\usepackage{empheq}
\usepackage{bm}
\usepackage{enumitem}
\usepackage{multirow}
\usepackage{makecell}
\usepackage{comment}
\usepackage{xspace}
\usetikzlibrary{positioning, calc,decorations.pathreplacing}

\usepackage{todonotes}
\usepackage{marginnote}
\usepackage[labelformat=simple]{subcaption}

\usepackage{xspace,prettyref}

\newcommand{\prob}[1]{ \mathbb{P}\left\{ #1 \right\} }

\newcommand{\abs}[1]{ \left| #1 \right| }

\newcommand{\Binom}{{\rm Binom}}

\newcommand{\eg}{e.g.\xspace}
\newcommand{\ie}{i.e.\xspace}

\newrefformat{eq}{(\ref{#1})}
\newrefformat{chap}{Chapter~\ref{#1}}
\newrefformat{sec}{Section~\ref{#1}}
\newrefformat{algo}{Algorithm~\ref{#1}}
\newrefformat{fig}{Figure~\ref{#1}}
\newrefformat{tab}{Table~\ref{#1}}
\newrefformat{rmk}{Remark~\ref{#1}}
\newrefformat{clm}{Claim~\ref{#1}}
\newrefformat{def}{Definition~\ref{#1}}
\newrefformat{cor}{Corollary~\ref{#1}}
\newrefformat{lmm}{Lemma~\ref{#1}}
\newrefformat{prop}{Proposition~\ref{#1}}
\newrefformat{app}{Appendix~\ref{#1}}
\newrefformat{hyp}{Hypothesis~\ref{#1}}
\newrefformat{thm}{Theorem~\ref{#1}}

\newtheorem{theorem}{Theorem}
\newtheorem{lemma}{Lemma}

\newcommand{\calG}{{\mathcal{G}}}
\newcommand{\ntest}{{n_{\rm test}}}
\newcommand{\ntrain}{{n_{\rm train}}}

\newcommand{\ER}{Erd\H{o}s-R\'enyi\xspace}

\newcommand{\mean}{\text{mean}}

\title{SeedGNN: Graph Neural Network for Supervised\\ Seeded Graph Matching}

\author{Liren Yu, Jiaming Xu, and Xiaojun Lin
\thanks{
L.\ Yu  and X.\ Lin  are with Elmore Family School of Electrical and Computer Engineering, Purdue University, 
West Lafayette, IN 47907, USA, \texttt{yu827@purdue.edu, linx@ecn.purdue.edu}.
J.\ Xu is with The Fuqua School of Business, Duke University, Durham,  NC 27708, USA, \texttt{jx77@duke.edu}.
}
}

\begin{document}

\maketitle

\begin{abstract}
There is a growing interest in designing Graph Neural Networks (GNNs) for seeded graph matching, which aims to match two unlabeled graphs using only topological information and a small set of seed nodes. However, most previous GNNs for this task use a semi-supervised approach, which requires a large number of seeds and cannot learn knowledge that is transferable to unseen graphs. In contrast, this paper proposes a new supervised approach that can learn from a training set how to match unseen graphs with only a few seeds. Our SeedGNN architecture incorporates several novel designs, inspired by theoretical studies of seeded graph matching: 1) it can learn to compute and use witness-like information from different hops, in a way that can be generalized to graphs of different sizes; 2) it can use easily-matched node-pairs as new seeds to improve the matching in subsequent layers. We evaluate SeedGNN on synthetic and real-world graphs and demonstrate significant performance improvements over both non-learning and learning algorithms in the existing literature. Furthermore, our experiments confirm that the knowledge learned by SeedGNN from training graphs can be generalized to test graphs of different sizes and categories.
\end{abstract}

\section{Introduction}\label{sec:intro}

Graph matching, also known as network alignment, aims to find the node correspondence between two graphs that maximally aligns their edge sets. 
As a ubiquitous but challenging problem, graph matching has numerous applications, including social network analysis \cite{narayanan2008robust,narayanan2009anonymizing,zafarani2015user,zhang2015cosnet,zhang2015multiple,chiasserini2016social},  computer vision \cite{conte2004thirty,schellewald2005probabilistic,vento2013graph}, natural language processing \cite{haghighi2005robust}, and computational biology \cite{singh2008global,kazemi2016proper,kriege2019chemical}.
This paper focuses on seeded graph matching, where a small portion of the node correspondence between the two graphs is revealed as seeds, and we seek to complete the correspondence using the few seeded node-pairs.
Seeded graph matching is motivated by the fact that, in many real applications, the correspondence between a small portion of the two node sets is naturally available. For example, in social network de-anonymization, some users who explicitly link their accounts across different social networks could become seeds ~\cite{narayanan2008robust,narayanan2009anonymizing}. 
Knowledge of even a few seeds has been shown to significantly improve the matching results for many real-world graphs \cite{kazemi2015growing,fishkind2019seeded}.

\begin{figure*}[ht]
\vspace{-8pt}
\centering
\begin{subfigure}[t]{0.36\textwidth}
\centering
\raisebox{1pt}{
\resizebox{0.7\textwidth}{!}{
\begin{tikzpicture} [scale = 1,
    node distance = 1.3cm, > = stealth, bend angle = 80, auto,
    seed/.style = {circle, draw = red!50, fill = red!30, thick, minimum size = 1mm,, inner sep=0pt},
    vertex1/.style = {circle, draw = black!50, thick, minimum size = 4.5mm},
    vertex2/.style = {circle, draw = black!50, thick, minimum size = 1.5mm, inner sep=0pt},
     graph/.style = {rectangle,rounded corners, draw = black!75,thick}
    ]
\centering

\node (G1) at (-2,0.6) [graph,text height = 1.2cm, minimum width = 2cm]{};
\node at (-2.7,1) {\large $\mathcal{G}_1$};
\node (G2) at (-2,-1) [graph,text height = 1.2cm, minimum width = 2cm]{};
\node at (-2.7,-0.6) {\large $\mathcal{G}_2$};

\node (n0) at (-2.7,0) [vertex2]{};
\node (n1) at (-2.5,0.5) [vertex2]{};
\node (n2) at (-2,1) [vertex2]{};
\node (n3) at (-2.2,0.2) [vertex2]{};
\node (n4) at (-1.5,0.6) [vertex2]{};
\node (n5) at (-1.8,0.2) [vertex2]{};

\node (pi0) at (-2.6,-1.5) [vertex2]{};
\node (pi1) at (-2.5,-1.1) [vertex2]{};
\node (pi2) at (-2,-0.6) [vertex2]{};
\node (pi3) at (-2.2,-1.4) [vertex2]{};
\node (pi4) at (-1.5,-1) [vertex2]{};
\node (pi5) at (-1.8,-1.4) [vertex2]{};

\draw  (n1) -- (n2);
\draw  (n1) -- (n3);
\draw  (n2) -- (n4);
\draw  (n3) -- (n5);
\draw  (n4) -- (n5);
\draw  (n1) -- (n4);
\draw  (n5) -- (-1.5,0.1);
\draw (n1) -- (-2.8,0.4);
\draw (n1) -- (n0);
\draw (n2) -- (-1.7,1.2);
\draw (n4) -- (-1.2,0.5);

\draw  (pi1) -- (pi2);
\draw  (pi2) -- (pi3);
\draw  (pi3) -- (pi4);
\draw  (pi2) -- (pi4);
\draw  (pi1) -- (pi5);
\draw  (pi2) -- (pi5);
\draw (pi1) -- (-2.8,-1.2);
\draw (pi3) -- (pi0);
\draw (pi2) -- (-1.7,-0.4);
\draw (pi4) -- (-1.3,-1.3);
\draw (pi4) -- (-1.2,-0.8);

\draw [thick,->] (G1) -- (0,0.6) ;
\draw [thick,->] (G2) -- (0,-1);
\node at(-0.55,0.85) {\small GNN};
\node at(-0.55,-0.735) {\small GNN};

\draw [->,thick] (0,0) -- (0,1.4);
\draw [->,thick] (0,0) -- (1.6,0);

\node (i1) at (0.7,1.2) [seed]{};
\node (i2) at (0.7,0.2) [seed]{};
\node at (0.55,1) {\Huge .};
\node at (0.65,0.8) {\Huge .};
\node at (0.3,0.75) {\Huge .};
\node at (0.7,0.5) {\Huge .};
\node at (0.4,0.4) {\Huge .};
\node at (0.5,0.5) {\Huge .};
\node at (0.55,0.3) {\Huge .};
\node at (0.4,0.7) {\Huge .};

\draw [->,thick] (0,-1.6) -- (0,-0.2);
\draw [->,thick] (0,-1.6) -- (1.6,-1.6);

\node (j1) at (0.7,-0.4) [seed]{};
\node (j2) at (0.7,-1.4) [seed]{};
\node at (0.5,-0.6) {\Huge .};
\node at (0.85,-0.7) {\Huge .};
\node at (1,-0.85) {\Huge .};
\node at (0.7,-1.1) {\Huge .};
\node at (0.8,-1.2) {\Huge .};
\node at (0.65,-0.75) {\Huge .};
\node at (0.55,-1.2) {\Huge .};
\node at (0.4,-1) {\Huge .};

\node [text=red] at (1.42,0.7) {seeds};
\draw [color=red] (i1) .. controls (1,0.4) .. (j1);
\draw [color=red] (i2) .. controls (1,-0.6) .. (j2);

\node at (0.8,-1.9){embedding space};

\end{tikzpicture}}}
\caption{In semi-supervised learning, only one pair of graphs $\mathcal{G}_1$ and $\mathcal{G}_2$ is provided. The GNN is applied to each graph separately. Taking the topological and non-topological features as input, this GNN maps each node to a node embedding in some space and matches nodes across two graphs based on the closeness in node embeddings. Using seeds as the training set, this GNN is trained so that the seed pairs have close node embeddings. However, when the number of seeds is small (while the GNN complexity is high), other true pairs may still have very different embeddings, leading to low matching accuracy.}
\label{fig:SSL-GNN}
\end{subfigure}
\hfill
\begin{subfigure}[t]{0.34\textwidth}
\centering
\includegraphics[width=0.97\textwidth]{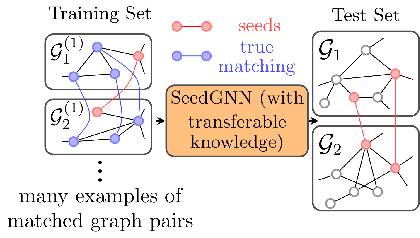}
\caption{Our SeedGNN is trained with many examples of matched graph pairs (on the left). Each training graph pair comes with the true matching (in blue), as well as the seeds (in red). With training, SeedGNN can distill transferable knowledge on how to effectively use a small number of seeds to best match other nodes. This SeedGNN is then applied to an unseen graph pair (on the right) to achieve high matching accuracy with only a limited number
of seeds.}
\label{fig:SeedGNN}
\end{subfigure}
\hfill
\begin{subfigure}[t]{0.26\textwidth}
\centering
\includegraphics[width=0.97\textwidth]{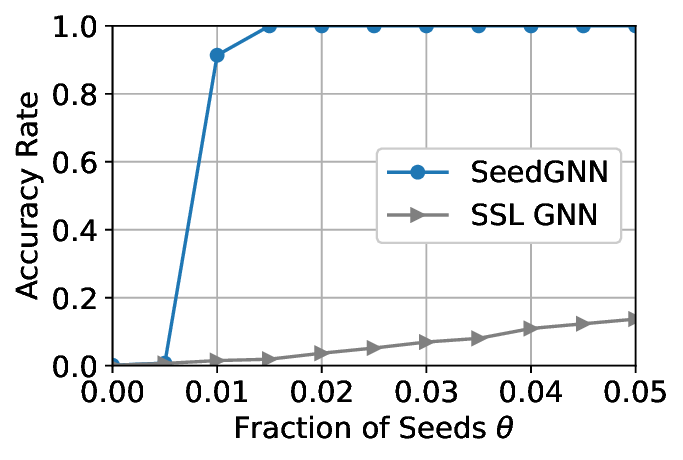}
\caption{Performance comparison on correlated \ER graphs shows that, with a very small fraction of seeds, SeedGNN can already achieve high matching accuracy, while SSL GNN almost completely fails.  See detailed set-up in \prettyref{sec:exp-set}.}
\label{fig:CompareER0}
\end{subfigure}
\caption{ Illustration of the advantages of supervised learning over semi-supervised learning for seeded graph matching.  
}\label{fig:illustration-seedgnn}
\end{figure*}

Recently, the Graph Neural Network (GNN) approach for graph matching has attracted much research attention.
Although such a machine-learning-based approach usually does not possess provable theoretical guarantees, it has the potential to learn valuable features from a large set of training data. Unfortunately, to date GNN has not been successfully applied to seeded graph matching.  
 Most previous GNNs for seeded graph matching are limited to a \emph{semi-supervised} learning (SSL) paradigm, which only operates on a \emph{single} pair of graphs  and treats the seed set as the labelled training data \cite{zhang2019graph,li2019partially,li2019adversarial,li2019graph,zhou2019disentangled,chen2020multi,derr2019deep}. The goal is to learn from the seed set useful features that can be used to compute node embeddings for all nodes (see \prettyref{fig:SSL-GNN}). However, this semi-supervised learning suffers from two major limitations. First, in order to obtain high matching accuracy, the set of seeds needs to be large, which is often unrealistic in practice. Second, as  this semi-supervised setting only learns \emph{within} a given pair of graphs, there is no effort in \emph{transferring knowledge} from one pair of graphs to other pairs of unseen graphs, which severely limits GNNs' potential in distilling the common knowledge from a large set of training graphs.
A natural but fundamental question is that

\emph{Can we learn to match two graphs with only a few seeds while generalizing to unseen graphs?}

This paper provides an affirmative answer to this question.
Specifically, we design a novel GNN architecture through a supervised approach, namely SeedGNN, that can learn from
many examples of matched graphs, distill the knowledge into the trained model automatically, and then
apply such knowledge to match unseen graph pairs with only a small number of seeds (see an illustration in \prettyref{fig:SeedGNN}). The performance comparison on correlated \ER graphs \cite{pedarsani2011privacy} shown in \prettyref{fig:CompareER0} demonstrates that our SeedGNN requires significantly fewer seeds to achieve higher matching accuracy than the SSL GNN.

\begin{figure}[ht]
  \centering
  \resizebox{0.6\columnwidth}{!}{
\begin{tikzpicture} [scale = 1,
    node distance = 1.3cm, > = stealth, bend angle = 80, auto,
    seed/.style = {circle, draw = red!50, fill = red!20, thick, minimum size = 4.5mm},
    vertex/.style = {circle, draw = blue!50, fill = blue!20, thick, minimum size = 4.5mm},
    vertex2/.style = {circle, draw = black!50, thick, minimum size = 4.5mm},
     graph/.style = {rectangle,rounded corners, draw = black!75,thick}
    ]
\centering
\node (n1) at (-1.7,0.45) [vertex2,label={[black]center:\small 1}]{};
\node (pi1) at (1.7,0.45) [vertex2,label={[black]center:\small $1'$}]{};
\node (n2) at (-0.6,1) [seed,label={[black]center:\small $2$}]{};
\node (pi2) at (0.6,1) [seed,label={[black]center:\small $2'$}]{};
\node (n3) at (-0.6,0.45) [vertex2,label={[black]center:\small $3$}]{};
\node (pi3) at (0.6,0.45) [vertex2,label={[black]center:\small $3'$}]{};

\draw [thick,color = red!70] (n2) -- (pi2);
\draw  (n1) -- (n2);
\draw  (n1) -- (n3);
\draw  (pi1) -- (pi2);
\draw  (pi1) -- (pi3);

\node (n4) at (-3.5,0.45) [vertex2,label={[black]center:\small 4}]{};
\node (pi4) at (3.5,0.45) [vertex2,label={[black]center:\small $4'$}]{};
\node (n5) at (-2.4,1) [seed,label={[black]center:\small $5$}]{};
\node (pi5) at (2.4,1) [seed,label={[black]center:\small $5'$}]{};
\node (n6) at (-2.4,0.45) [vertex2,label={[black]center:\small $6$}]{};
\node (pi6) at (2.4,0.45) [vertex2,label={[black]center:\small $6'$}]{};

\draw [thick, color = red!70] (n5) .. controls (-2,1.5) and (2,1.5) .. (pi5);
\draw  (n4) -- (n5);
\draw  (n4) -- (n6);
\draw  (pi4) -- (pi5);
\draw  (pi4) -- (pi6);

\node (G1) at (-2.1,0.73) [graph,text height = 0.95cm, minimum width = 4cm]{};
\node at (-3.7,1) {\large $\mathcal{G}_1$};
\node (G2) at (2.1,0.73) [graph,text height = 0.95cm, minimum width = 4cm]{};
\node at (3.7,1) {\large $\mathcal{G}_2$};
\end{tikzpicture}
}
\caption{ 
The node-pairs $(i,i')$ are true matches and the red node-pairs  are seeds. 
}
\label{fig:graph-example}
\end{figure}
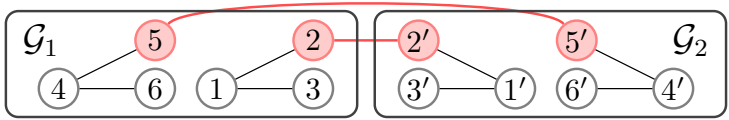

Our supervised SeedGNN incorporates two valuable insights from theoretical studies of seeded graph matching.  The first insight is the notion of ``witnesses''~\cite{korula2014efficient,mossel2019seeded}. For node $v$ in graph $\calG_1$ and node $w$ in graph $\calG_2$, we say a seed pair $(u,u')$ is a \emph{$\ell$-hop witness} for $(v,w)$ if $u$ is an $\ell$-hop neighbor of $v$ in $G_1$ and $u'$ is an $\ell$-hop neighbor of $w$ in $G_2$. For example, in \prettyref{fig:graph-example} the seed pair $(2,2')$ is a 1-hop witness for $(1,1')$.
For many graph matching problems, a true pair tends to have more witnesses than fake pairs \cite{korula2014efficient,mossel2019seeded}. Thus, the number of witnesses can be used to measure the similarity of node-pairs.
The second insight is the notion of ``percolation.'' In particular, note that some true pairs are easier to match than others. Some theoretical algorithms use these easily-matched true pairs as new seeds to match other nodes \cite{korula2014efficient,kazemi2015growing}. For example, in \prettyref{fig:graph-example}, since the node-pairs $(1,1')$ and $(4,4')$ have a 1-hop witness, these two node-pairs can be matched first. Then, used as new seeds, they become the witnesses for other node-pairs, such as $(3,3')$ and $(6,6')$.
When done properly, a percolation process can be triggered to match a large number of nodes. 
For some graphs, these two insights have been crucial for theoretical algorithms to successfully match graphs of $n$ nodes with only $\Theta(\log n)$ seeds \cite{korula2014efficient,kazemi2015growing,yu2021power}.


Compared to the existing GNNs, our SeedGNN architecture is carefully designed to effectively and explicitly incorporate the above two insights.
First, most existing GNNs are ``node-based,'' i.e., they are applied to each node to compute a node embedding. In contrast, SeedGNN is ``pair-wise,'' i.e., it is applied to each pair of nodes across the two graphs. As we elaborate in \prettyref{sec:convolution}, this pair-wise architecture is much more effective in learning how to compute and use witnesses in a way that can be generalized to graphs with different sizes.
Second, we carefully design a percolation module to filter out node-pairs with low similarities.  As a result, only the ``cleaner'' new seeds are used to trigger the percolation process. 

Note that a similar pair-wise architecture has appeared before in  NGM \cite{wang2021neural} for seedless graph matching.  However, the NGM architecture was not designed for seeded graph matching, and there are crucial differences that prevent the effective use of the witnesses and percolation ideas. We analytically show in \prettyref{sec:analysis} that, for seeded graph matching, such differences can significantly affect the matching performance when the number of seeds is small. Our experiments in \prettyref{sec:exp-res} further verify that NGM does not generalize well when the test graphs have much larger sizes and node-degrees than the training graphs.

Our numerical experiments (in \prettyref{sec:experiment}) on both synthetic and real-world graphs show that SeedGNN
significantly outperforms the state-of-the-art algorithms, including both non-learning and learning-based ones,
in terms of seed-size requirement and matching accuracy. Moreover, our SeedGNN can generalize to 
match unseen graphs of sizes and types  different from the training set.

\section{Further Related Work}\label{sec:related-work}

In this section, we discuss further related work.  More discussion on additional related work is deferred to \prettyref{app:additional-related-work}.

\vspace{-3pt}
\paragraph{Theoretical Algorithms}
Various seeded matching algorithms have been proposed based on hand-designed similarity metrics computed from local topological structures \cite{pedarsani2011privacy,yartseva2013performance,korula2014efficient,shirani2017seeded,mossel2019seeded,yu2021power}. The theoretical analysis on these algorithms explains why a particular set of features (e.g., witnesses \cite{korula2014efficient} and percolation \cite{yartseva2013performance}) are valuable for graph matching. However, these theoretical algorithms require carefully hand-tuned parameters and may not synthesize different features most effectively (see detailed discussion in Appendix \ref{app:additional-related-work}).
In contrast, SeedGNN 
can potentially learn (from the training data) what combinations of features are most useful, and thus outperform theoretical algorithms, as shown in our experiments \prettyref{sec:exp-res}. 

\paragraph{GNN for Seedless Graph Matching}
As aforementioned, most existing GNNs for seeded graph matching take a semi-supervised learning approach. In contrast, our SeedGNN falls into a supervised learning approach, which aims to transfer knowledge from training graphs to unseen graphs. In the literature, such a supervised learning approach has been applied to \emph{seedless} versions of the graph matching problems in \cite{zanfir2018deep,wang2019learning,wang2021neural,wang2020combinatorial,wang2021neural,jiang2019glmnet,wang2020learning,fey2020deep,rolinek2020deep,gao2021deep,pmlr-v139-yu21d}.
For such seedless matching problems, non-topological node features are often assumed to be available and informative.
Thus, a node-based GNN is effective in learning how to extract useful node representations from high-quality non-topological node features. 
However, for seeded matching problems, it is difficult to design a node-based GNN to effectively utilize seed information (see further discussions in \prettyref{sec:convolution}). In contrast, our pair-wise SeedGNN architecture is much more effective in learning how to use seed information.

\section{Problem Definition}\label{sec:problem}

We represent a graph of $n$ nodes by $\mathcal{G} = (V,\mathbf{A})$, where $V = \{1,2,...,n\}$ denotes the node set, and $\mathbf{A}\in \{0,1\}^{n\times n}$ denotes the adjacent matrix, such that $\mathbf{A}(i,j)=1$ if and only if nodes $i$ and $j$ are connected. 
For seeded graph matching, we are given two graphs $\mathcal{G}_1 = (V_1, \mathbf{A}_1)$ of $n_1$ nodes and $\mathcal{G}_2=(V_2,\mathbf{A}_2)$ of $n_2$ nodes. Without loss of generality, we assume $n_1\le n_2$.
There is an \emph{unknown} injective mapping $\pi:V_1\to V_2$ between $\mathcal{G}_1$ and $\mathcal{G}_2$. When $\pi(i)=j$, we say that $ i \in V_1$ corresponds to $j\in V_2$. Throughout the paper, we denote a node-pair by $(i,j)$, where $i\in V_1$ and $j\in V_2$. For each node-pair $(i,j)$, if $j = \pi(i)$, then $(i,j)$ is a \emph{true pair}; if $j \neq \pi(i)$, then $(i,j)$ is a \emph{fake pair}. Then, a seed set $\mathcal{S}$ containing a fraction of true pairs is given.
The goal of seeded graph matching is to recover the ground-truth mapping $\pi$ based on the observation of $\mathcal{G}_1$, $\mathcal{G}_2$ and $\mathcal{S}$. 

In this work, we consider the problem of seeded graph matching in the supervised setting. The training set consists of several pairs of graphs, their initial seeds, and ground-truth mappings. Specifically, we use $\mathcal{T}=\{({P}^{(1)},\pi^{(1)}),({P}^{(2)},\pi^{(2)}),...,({P}^{(N)},\pi^{(N)})\}$ to denote the training set, where ${P}^{(i)}=(\mathcal{G}_1^{(i)},\mathcal{G}_2^{(i)},\mathcal{S}^{(i)})$ denotes the $i$-th training example and $\pi^{(i)}$ is the ground-truth mapping for the $i$-th training example.  For different training examples, the sizes of graphs and seed sets could be different. Our goal is to design  a GNN architecture that can learn from training examples to predict the ground-truth mappings for unseen test graphs.

\section{The Proposed Method}\label{sec:method}

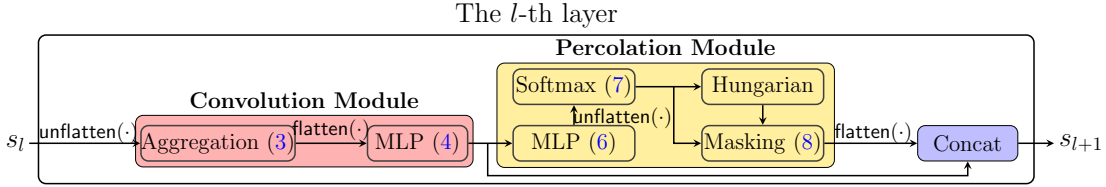
\begin{figure*}[ht]
\centering
\resizebox{0.9\textwidth}{!}{
\begin{tikzpicture} [scale = 1,
    node distance = 1.3cm, > = stealth, bend angle = 80, auto,
    GNNlayer/.style = {rectangle, rounded corners,draw = black!75,thick},
    dot/.style={circle, fill=black, inner sep=0.7pt, outer sep=0pt },
    seed/.style = {circle, draw = red!50, fill = red!20, thick, minimum size = 4.5mm},
    vertex1/.style = {circle, draw = black!50, thick, minimum size = 4.5mm},
    vertex2/.style = {circle, draw = blue!50, fill = blue!20, thick, minimum size = 4.5mm},
     graph/.style = {rectangle,rounded corners, draw = black!75,thick}
    ]

\node (GNNl) at (-4.2,0.63) [rectangle,rounded corners,text height = 2.5cm, text width = 18.2cm, draw = black,thick, label={above:\Large The $l$-th layer}]{};

\node (convolution) at (-8.5,0.04) [rectangle,rounded corners,fill=red!30, text height = 0.7cm, text width = 6cm, draw = black, label={[black]above:\large \textbf{Convolution Module}}]{};
\node (propagation) at (-10.1,0) [rectangle,rounded corners,text height = 0.4cm, text width = 2.6cm, draw = black!75,thick,label={center:\large Aggregation \prettyref{eq:matrix-mul}}]{};
\node (t1) at (-6.4,0) [rectangle,rounded corners,text height = 0.4cm, text width = 1.6cm, draw = black!75,thick,label=center:\large MLP \prettyref{eq:NN-hl}]{};
\draw [->,thick] (propagation) -- (t1);
\node (vec) at (-8,0.25) {$\mathsf{flatten}(\cdot)$};

\node (percolation) at (-1.8,0.52) [rectangle,rounded corners,fill={rgb:orange,1;yellow,2;white,5},text height = 1.68cm, text width = 6cm, draw = black, label={[black]above:\large \textbf{Percolation Module}}]{};
\node (t2) at (-3.5,0) [rectangle,rounded corners,text height = 0.4cm, text width = 2cm, draw = black!75,thick,label=center:\large MLP \prettyref{eq:NN-rhol}]{};
\node (softmax) at (-3.5,1.05) [rectangle,rounded corners,text height = 0.4cm, text width = 2cm, draw = black!75,thick,label=center:\large Softmax \prettyref{eq:normalize-softmax}]{};
\node (hungarian) at (0,1.05) [rectangle,rounded corners,text height = 0.4cm, text width = 2cm, draw = black!75,thick,label=center:\large Hungarian]{};
\node (prod) at (0,0) [rectangle,rounded corners,text height = 0.4cm, text width = 2cm, draw = black!75,thick,label=center:\large Masking \prettyref{eq:mask}]{};
\node at (-2.6,0.47) {$\mathsf{unflatten}(\cdot)$};
\node at (2.05,0.2) {$\mathsf{flatten}(\cdot)$};

\draw [->,thick] (t2) -- (softmax);
\draw [->,thick] (softmax) -- (hungarian);
\draw [->,thick] (hungarian)  -- (prod);
\draw [->,thick] (-1.65,1.05) -- (-1.65,0) -- (prod);

\node (concat) at (3.8,0) [rectangle,rounded corners,fill={rgb:blue,1;white,3},text height = 0.4cm, text width = 1.6cm, draw = black,label=center:\large Concat]{};

\draw [->,thick] (prod) -- (concat);
\draw [->,thick] (t1) -- (t2);
\draw [->,thick] (-5.1,0) -- (-5.1,-0.6) -- (3.8,-0.6) -- (concat);

\node (s1) at (-13.85,0) {\Large $s_l$};
\draw [->,thick] (-13.6,0) -- (propagation);
\node at (-12.5,0.2) {$\mathsf{unflatten}(\cdot)$};

\node (s2) at (5.9,0) {\Large $s_{l+1}$};
\draw [->,thick] (concat) -- (5.4,0);

\end{tikzpicture}}
\caption{An overview of the $l$-th layer of our SeedGNN architecture. There are  $L$ layers in total and each layer consists of two main modules. With the node-pair representations $s_l$ as input, the convolution module is a \emph{local} processing step that aggregates the neighborhood information of each node-pair and updates the representation of its similarity through a neural network. The percolation module is a \emph{global} processing step that compares the updated similarities of all node-pairs and finds the high-confidence ones. Then, we combine the local and global information from the two modules and propagate the new representations $s_{l+1}$ to the next layer.}
\label{fig:Architecture}
\end{figure*}

In this section, we present in detail our SeedGNN for seeded graph matching.  See \prettyref{fig:Architecture}  for a high-level illustration. 

\vspace{-10pt}
\paragraph{Notation} we use $\mathsf{flatten}(\cdot)$ to denote the matrix reshape operation that converts a $n_1\times n_2\times d$ matrix to a matrix of $n_1n_2\times d$, where the $(i,j,:)$-th entry of the input matrix is the $((i-1)n_2 +j,:)$-th entry of the output matrix. Then, we use  $\mathsf{unflatten}(\cdot)$ to denote the inverse operation of $\mathsf{flatten}(\cdot)$.
\vspace{-10pt}


\subsection{Generalizable Encoding Method for Seeds}\label{sec:seedencode}
We encode the seeded relationship as inputs for our SeedGNN. More precisely, let $s_1\in \{0,1\}^{n_1n_2\times 1}$ be the indicator vector for seeds among $n_1n_2$ node-pairs.
If the node-pair $(i,j)$ is a seed, we let the $((i-1)n_2+j)$-th entry of $s_1$ be 1, and  0 otherwise.  

We contrast our way of encoding seeds with an alternate one-hot encoding method. One-hot encoding assigns the $i$-th seeded node with a binary vector, whose $i$-th element being 1, and all other elements are 0.
The benefit of our encoding method is that  the dimension of the encoding vector is fixed at 1 for each node-pair, which does not depend on the graph size or the number of seeds. Thus, SeedGNN with our encoding method can be applied to unseen graphs with arbitrary graph sizes and numbers of seeds. In contrast, one-hot encoding method needs to pre-specify the maximum number of seeds, and thus GNNs with one-hot encoding can not generalize to new graphs with even more seeds. 


\subsection{Convolution Module}\label{sec:convolution}



With the seed information encoded as 0/1 for each node-pair,  we still need to carefully design a GNN architecture that can count witnesses. 
Note that most existing GNN approaches for graph matching are ``node-based'' \cite{zhang2019graph,chen2020multi,wang2019learning,wang2020combinatorial,fey2020deep,rolinek2020deep}. They apply a common GNN separately to each of the two graphs in order to learn a node embedding \emph{for each node}. They then match nodes in the two graphs based on the similarity of the corresponding node embeddings. 
However, it is difficult for these approaches to utilize our newly-encoded seed information effectively.
As shown in \prettyref{fig:graph-example}, our encoding of seed information can also be viewed as ``cross-links" (highlighted in red color) across the two graphs. With these ``cross-links'', we can then combine the two graphs together and apply the node-based GNN on this union graph. However, the topological structure of this union graph only informs the GNN that there is a seed at a particular location in the neighborhood, but not the seed identity. 
For example, in \prettyref{fig:graph-example}, even though node $1'$ and node $4'$ have different seeds in their neighborhoods, their local neighborhood topologies (and the seed positions) look exactly the same. 
Thus, node-based GNN will have a hard time to come up with node embeddings such that node 1 has a close embedding to node $1'$ but not close to node $4'$. Instead, our SeedGNN is ``pair-wise", i.e., it is applied on node-pairs instead of nodes. Intuitively, when we apply such a pair-wise GNN to the node-pairs $(1,1')$ and $(1,4')$ in \prettyref{fig:graph-example}, it can easily tell that $(1,1')$ has a witness, while $(1,4')$ does not. As a result, this pair-wise GNN will count and utilize witnesses easily.


Specifically, taking the seed encoding vector $s_1$ as input, the counting of 1-hop witnesses can be written as
\vspace{-1pt}
\begin{align}\label{eq:1-hop}
h_1= (\mathbf{A}_1\otimes \mathbf{A}_2)s_1
\vspace{-1pt}
\end{align}
where $\otimes$ denotes the Kronecker product. Applying \prettyref{eq:1-hop} to \prettyref{fig:graph-example}, we can get that the node-pairs $(1,1')$ and $(4,4')$ have a 1-hop witness, respectively.
Likewise, we may further compute the $l$-hop witness-like information $h_l$ in the $l$-th layer of our SeedGNN as
\vspace{-1pt}
\begin{align}
    h_{l}=(\mathbf{A}_1\otimes \mathbf{A}_2)s_l, \label{eq:kron-progagation}
\end{align}
\vspace{-1pt}
where $s_l\in \mathbb{R}^{n_1n_2\times d_l}$ is specified later in Section~\ref{sec:percolation}, which contains the witness-like information within $(l-1)$-hops. 
Note that \prettyref{eq:kron-progagation} can be expanded as,  for node-pair $(i,j)$, 
$$
h_l[(i-1)n_2+j,:]=\sum_{\substack{(u,v):\\  \mathbf{A}_1(u,i)= \mathbf{A}_2(v,j)=1}} s_l[(u-1)n_2+v,:],
$$
which is similar to the aggregation step of the standard GNN in \cite{hamilton2017inductive}. The only difference is that we aggregate over a node-pair's neighborhood.
A direct implementation of \prettyref{eq:kron-progagation} takes $O(n_1^2n_2^2)$ computation, but we can reduce the complexity by letting $H_{l}=\mathsf{unflatten}(h_{l})$ and $S_l= \mathsf{unflatten}(s_l)$, and rewriting \prettyref{eq:kron-progagation} as
\begin{align}
  H_{l}[:,:,t]=& \mathsf{unflatten}((\mathbf{A}_1\otimes \mathbf{A}_2)s_l[:,t])\nonumber
  \\=&\mathbf{A}_1 S_l[:,:,t] \mathbf{A}_2,\qquad t =1,2,...,d_l. \label{eq:matrix-mul}
  \vspace{-1pt}
\end{align}
Assume that the mean of the node degrees of $\mathcal{G}_1$ and $\mathcal{G}_2$ is at most $d_{\mean}$. When we represent $A_1$ and $A_2$ with sparse matrices, each of them only contain $n_1d_{\text{mean}}$ and $n_2d_{\text{mean}}$ elements. Thus, by sparse matrix multiplication, the time complexity of Equation (3) is $O(n_1n_2d_{\text{mean}})$.

As we will see later in \prettyref{sec:percolation}, $s_l$ will also contain outputs from the percolation layer. In order to learn how to best synthesize these two features, we apply a neural network on $h_l$ after \prettyref{eq:kron-progagation}:
\vspace{-1pt}
\begin{align}
    m_l =\phi_l(h_l),\label{eq:NN-hl}
\end{align}
where the update function $\phi_l$ is implemented as a $K$-layer neural network (we use $K=2$ in our experiment). Let $\phi_l^{[0]}(h_l)=h_l$. The $k$-th layer of $\phi_l$ can be formulated as
\begin{align}
    \phi_l^{[k]}(h_l) =\sigma\left(\phi_l^{[k-1]}(h_l)\bm{W}^{[k-1]}+\bm{b}^{[k-1]}\right),\label{eq:neuralnet}
\end{align}
where $\bm{W}^{[k-1]}$ and $\bm{b}^{[k-1]}$ are learnable weights, initialized as Gaussian random variables; $\sigma$ is an activation function (we use ReLU).  
The updated representations $m_l\in \mathbb{R}^{n_1n_2\times (d_{l}-1)}$ will be sent to the next module of SeedGNN.



\begin{figure*}[ht]
\centering
\begin{subfigure}[t]{0.3\textwidth}
\centering
\includegraphics[width=0.9\columnwidth]{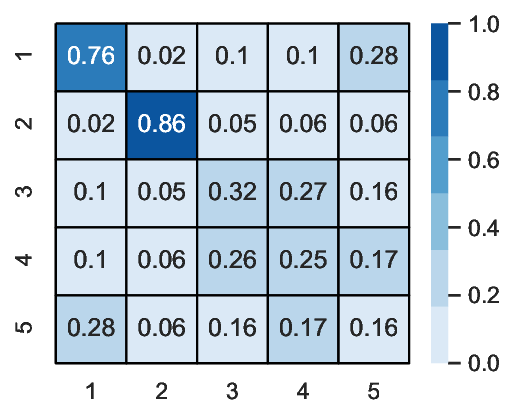}
\caption{Similarity/confidence matrix $Y_l$.}
\label{fig:softcorr}
\end{subfigure}
\quad
\begin{subfigure}[t]{0.3\textwidth}
\centering
\includegraphics[width=0.9\columnwidth]{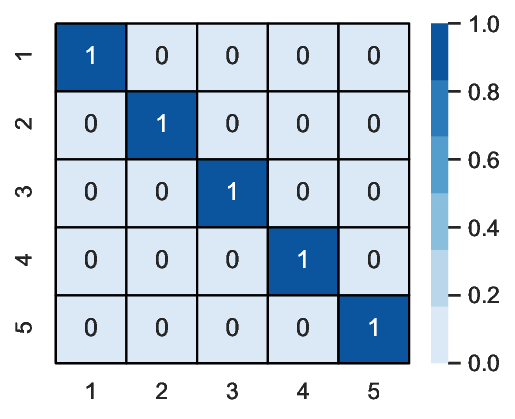}
\caption{Output $R_l$ of the Hungarian algorithm.}
\label{fig:hun}
\end{subfigure}
\quad
\begin{subfigure}[t]{0.3\textwidth}
\centering
\includegraphics[width=0.9\columnwidth]{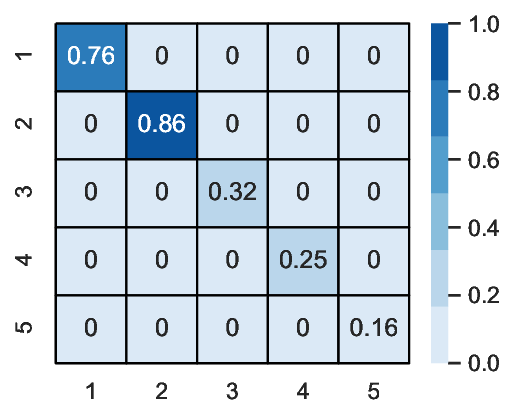}
\caption{$Z_l$ produced by masking.}
\label{fig:mask}
\end{subfigure}
\caption{Illustration of percolation module. The diagonal (resp.\ off-diagonal) entries correspond to true (resp.\ fake) pairs.
}
\end{figure*}

\subsection{Percolation Module}\label{sec:percolation}

The percolation module is designed to match high-confidence nodes at one layer and to propagate the matched nodes as new seeds to the subsequent layers. 
Formally, we first obtain a similarity matrix in the $l$-th layer by mapping the node-pair representations $m_l$ to a 1-dimension vector, which is used to assess the similarity of each node-pair:
\begin{align}
\vspace{-1pt}
   x_l = \rho_l(m_l).\label{eq:NN-rhol}
   \vspace{-1pt}
\end{align}
We implement $\rho_l$ as a multi-layer neural network that is defined similarly as $\phi_l$ in \prettyref{eq:neuralnet}. The output $x_l$ is in $\mathbb{R}^{n_1n_2\times 1}$.
Then, we transform $x_l$ to $X_l=\mathsf{unflatten}(x_l)\in \mathbb{R}^{n_1\times n_2}$, and apply row-wise softmax to normalize $X_l$ and obtain the similarity (confidence) matrix $Y_l$ for node-pairs:
\begin{align}
\vspace{-1pt}
    Y_l = \left(\text{softmax}(X_l)+\text{softmax}(X_l^\top)^\top\right)/2,\label{eq:normalize-softmax}
\end{align}
where for each row $\bm{v}=(v_1,...,v_n)\in\mathbb{R}^n$ of input matrix, the softmax function is defined as
\begin{align*}
   \text{softmax}(\bm{v})_i=\frac{\exp(v_i)}{\sum_{j=1}^{n}\exp(v_j)},\quad \text{for } i = 1,2,..,n.
   \vspace{-1pt}
\end{align*}
The similarity matrix $Y_l$ needs ``cleaning" because it contains a lot of ``noisy" information. For example, many fake pairs may possess comparable similarity with true pairs (see \prettyref{fig:softcorr} for example). Further, there are far more fake pairs than true pairs. As a result, directly utilizing such misleading information may lead to even more matching errors. Inspired by the percolation idea from theoretical algorithms, which passes only new seeds with high confidence levels to the next stage \cite{yartseva2013performance}, we leverage an approach called ``masking" to remove the noisy information and retain the cleaner information in $Y_l$. Specifically, we utilize the Hungarian matching algorithm \cite{edmonds1972theoretical} to solve a linear assignment problem on $Y_l$ to find an injective mapping between $\mathcal{G}_1$ and $\mathcal{G}_2$, such that the total similarity of the matched node-pairs is maximized (see \prettyref{fig:hun} for example). The matching result is denoted by $R_l\in\{0,1\}^{n_1\times n_2}$, where $R_l(i,j)=1$ if the node-pair $(i,j)$ is matched by the Hungarian algorithm, and $R_l(i,j)=0$ otherwise. Then, we filter out the noisy information in $Y_l$ by ``masking":
\begin{align}
\vspace{-8pt}
    z_l=\mathsf{flatten}( Y_l\circ R_l),\label{eq:mask}
\vspace{-2pt}
\end{align}
where $\circ$ denotes element-wise multiplication (see \prettyref{fig:mask} for example). The matching information $z_l$ is sent to the next layer. As a result, many noisy node-pairs are discarded. We note that both the idea of using similarity matrix to refine higher-layer matching and the idea of masking have appeared in seedless matching \cite{wang2019learning,fey2020deep,yu2019learning}. However, \cite{wang2019learning,fey2020deep}  do not clean up the ``noisy'' information as we carefully did, and \cite{yu2019learning} only applies the Hungarian algorithm in their loss function (but not the intermediate layers). Readers can refer to the numerical results in \prettyref{app:exp-design}, which demonstrate the importance of carefully cleaning up ``noisy'' information in each layer.
Further, unlike previous percolation algorithms \cite{yartseva2013performance}, our design of the percolation module can correct matching errors from earlier layers.
We discuss these differences further in Appendix \ref{app:additional-related-work}.

\paragraph{The combination of the two features}
With the convolution module and the percolation module, our SeedGNN can identify witnesses-like information at different hops and generate new seeds for percolation. However, these capabilities alone are insufficient. For example, when graphs are very sparse, even true node-pairs may not have enough witnesses if the number of hops $l$ is small. When graphs are very dense, a fake pair may also have many witnesses if $l$ is large. Thus, SeedGNN needs to learn how to adaptively utilize various types of witnesses in different types of graphs.   Similarly, even with the above ``cleaning" procedure, the output of the percolation module may still have low-confidence seeds. Directly using them for percolation could lead to cascading errors. Thus, SeedGNN also needs to learn how to use new seeds with different levels of confidence.

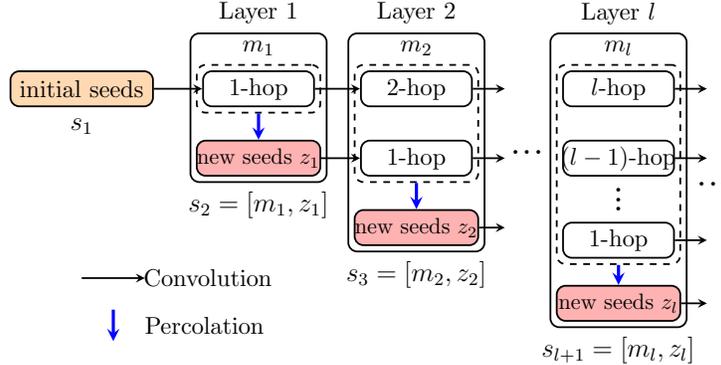
\begin{figure}[ht]
    \centering
    \resizebox{0.6\textwidth}{!}{
\begin{tikzpicture} [scale = 1,
    node distance = 1.3cm, > = stealth, bend angle = 80, auto,
    GNNlayer/.style = {rectangle, rounded corners,draw = black!75,thick},
    dot/.style={circle, fill=black, inner sep=0.7pt, outer sep=0pt }
    ]

\node (layer1) at (-4.7,1.2) {Layer 1};
\node (layer2) at (-2.2,1.2) {Layer 2};
\node (layerl) at (1,1.2) {Layer $l$};

\node (S1) at (-7.5,0) [rectangle,rounded corners,text height = 0.3cm, text width = 2cm, draw = black,thick,fill=orange!30, label={center:\shortstack{initial seeds}}]{};
\node at (-7.5,-0.6) {\large $s_1$};
\node (layer1-1hop) at (-4.7,0) [rectangle,rounded corners,text height = 0.3cm, text width = 1.5cm, draw = black,thick, label={center:$1$-hop}]{};
\node (Z1) at (-4.7,-1.1) [rectangle,rounded corners,text height = 0.3cm, text width = 1.7cm, draw = black,thick,fill=red!30, label={center:\small new seeds $z_1$}]{};
\node (m1) at (-4.7,0) [rectangle,rounded corners,text height = 0.5cm, text width = 1.7cm, draw = black,thick, dashed,label={above:$m_1$}]{};
\node (s1) at (-4.7,-0.3) [rectangle,rounded corners,text height = 2.1cm, text width = 1.9cm, draw = black,thick,label={below:\large $s_2=[m_1,z_1]$}]{};
\node (layer2-2hop) at (-2.2,0) [rectangle,rounded corners,text height = 0.3cm, text width = 1.5cm, draw = black,thick, label={center:$2$-hop}]{};
\node (layer2-1hop) at (-2.2,-1.1) [rectangle,rounded corners,text height = 0.3cm, text width = 1.5cm, draw = black,thick, label={center:$1$-hop}]{};
\node (Z2) at (-2.2,-2.2) [rectangle,rounded corners,text height = 0.3cm, text width = 1.7cm, draw = black,thick,fill=red!30, label={center:\shortstack{\small new seeds $z_2$}}]{};
\node (m2) at (-2.2,-0.55) [rectangle,rounded corners,text height = 1.6cm, text width = 1.7cm, draw = black,thick, dashed,label={above:$m_2$}]{};
\node (s2) at (-2.2,-0.85) [rectangle,rounded corners,text height = 3.2cm, text width = 1.9cm, draw = black,thick,label={below:\large $s_3=[m_2,z_2]$}]{};
\node (layerl-lhop) at (1,0) [rectangle,rounded corners,text height = 0.3cm, text width = 1.5cm, draw = black,thick, label={center:$l$-hop}]{};
\node (layerl-2hop) at (1,-1.1) [rectangle,rounded corners,text height = 0.3cm, text width = 1.5cm, draw = black,thick, label={center:$(l-1)$-hop}]{};
\node (layerl-1hop) at (1,-2.4) [rectangle,rounded corners,text height = 0.3cm, text width = 1.5cm, draw = black,thick, label={center:$1$-hop}]{};
\node (ml) at (1,-1.2) [rectangle,rounded corners,text height = 2.9cm, text width = 1.7cm, draw = black,thick, dashed,label={above:$m_l$}]{};
\node (Zl) at (1,-3.4) [rectangle,rounded corners,text height = 0.3cm, text width = 1.7cm, draw = black,thick, fill=red!30,label={center:\shortstack{\small new seeds $z_l$}}]{};
\node (sl1) at (1,-1.45) [rectangle,rounded corners,text height = 4.4cm, text width = 1.9cm, draw = black,thick,label={below: \large $s_{l+1}=[m_l,z_l]$}]{};
\node at (-0.45,-1) {\huge ...};
\node at (2.5,-1.5) {\huge ...};
\draw [->,thick] (-7.5,-3) -- (-6.5,-3);
\node at (-5.5,-3) {Convolution};
\draw [->,line width = 0.5mm,blue] (-7,-3.5) -- (-7,-4);
\node at (-5.55,-3.75) {Percolation};

\draw [->,thick] (S1) -- (layer1-1hop);
\draw [->,thick] (layer1-1hop) -- (layer2-2hop);
\draw [->,thick] (Z1) -- (layer2-1hop);
\draw [->,thick] (layer2-2hop) -- (-0.8,0);
\draw [->,thick] (layer2-1hop) -- (-0.8,-1.1);
\draw [->,thick] (Z2) -- (-0.8,-2.2);
\draw [->,thick] (layerl-lhop) -- (2.4,0);
\draw [->,thick] (layerl-2hop) -- (2.4,-1.1);
\draw [->,thick] (layerl-1hop) -- (2.4,-2.4);
\draw [->,thick] (Zl) -- (2.4,-3.4);

\node at(1,-1.6) {\huge .};
\node at(1,-1.75) {\huge .};
\node at(1,-1.9) {\huge .};
\draw [line width = 0.5mm,->,blue] (m1) -- (Z1);
\draw [line width = 0.5mm,->,blue] (m2) -- (Z2);
\draw [line width = 0.5mm,->,blue] (ml) -- (Zl);
\end{tikzpicture}
}
\caption{The witness-like information and new seeds computed by each layer.}
    \label{fig:layerwise-feature}
\end{figure}
The neural module in \prettyref{eq:NN-hl} is precisely designed to enable such learning. Specifically, instead of directly using the output $z_l$ from the percolation module as new seeds, we concatenate it with the output of the convolution module, i.e., $s_{l+1}=[m_{l},z_{l}]\in \mathbb{R}^{n_1n_2\times d_{l+1}}$, as the input to the next layer. Then, after passing $s_{l+1}$ through \prettyref{eq:kron-progagation}, we apply the neural module \prettyref{eq:NN-hl}. The joint effect of this design is that SeedGNN can utilize the confidence levels of $z_l$ to decide how much it should rely on various types of witnesses. Intuitively, at a higher layer $l \ge 2$, after passing $s_l$ by \prettyref{eq:kron-progagation} and \prettyref{eq:NN-hl}, $m_l$ may contain $l$-hop witness-like information from the initial seeds $s_1$, $(l-1)$-hop witness-like information from new seeds $z_1$, ... and 1-hop witnesses information from new seeds $z_{l-1}$ (see \prettyref{fig:layerwise-feature}). However, unlike the initial seeds $s_1$ that are either 0 or 1, the new seeds $z_1,$ $z_2,$ $...,$ $z_{l-1}$ also come with confidence levels. Thus, thanks to the non-linearity in $\phi_l$ at each layer, the strength of the various types of witness-like information (from either the initial seeds or the new seeds) will vary depending on the confidence levels of the new seeds, which then  allows SeedGNN to learn how to best utilize them adaptively. 
For example, for sparse graphs, the confidence levels of the new seeds in the first several layers are low. As a result, SeedGNN can utilize witnesses based on the initial seeds but at a larger number of hops. In contrast, for dense graphs, if the confidence levels of the new seeds in the first several layers are already high, SeedGNN can then utilize the new witnesses computed from those new seeds. This capability is experimentally validated in Appendix~\ref{sec:layer-wise} by studying the layer-wise matching process of SeedGNN for different types of graphs.
Further, SeedGNN can even combine different types of witness-like information together and  extract more valuable features.

\subsection{Loss Function}\label{sec:loss}
Finally, we utilize the ground-truth node correspondence as the supervised training information for end-to-end training. More precisely, for any training example  $(P,\pi)\in \mathcal{T}$, we adopt the cross-entropy loss to measure the difference between our prediction and the ground-truth mapping $\pi$. Then, we add up the cross-entropy loss of every layer:
\begin{align*}
    \mathcal{L}_P(\vartheta) =  -\sum_{l=1}^{L}\left(\sum_{(i,j):\ j=\pi(i)}\log \left(Y_l(i,j)+\epsilon\right)+\sum_{(i,j),\ j\neq\pi(i)}\log \left(1-Y_l(i,j)+\epsilon\right)\right),
\end{align*}
where $Y_l$ is given in \prettyref{eq:normalize-softmax}, $\vartheta$ denotes all the learnable weights in the networks $\phi_l$ and $\rho_l$, and $\epsilon$ is a small positive value (e.g. $\epsilon=10^{-9}$) to avoid a logarithm of zero. 
The total loss function is $
\mathcal{L}(\vartheta) =  \sum_{P\in \mathcal{T}} \mathcal{L}_P(\vartheta).
$
We find that the use of the losses from all layers in training helps to speed up the training process. This is somewhat inspired by hierarchical learning methods in \cite{bengio2009learning,schmidhuber1992learning,Simonyan2015VeryDC}. It allows the lower layers to be trained first, making it easier to train the next layers. 
In testing, we will apply the trained SeedGNN model on the test graphs and only use the matching result of the final layer, $R_L$, as the predicted mapping since the final layer already synthesizes all the features learned at the lower layers. 

The total time complexity of SeedGNN is $O(n_1n_2^2)$, and the space complexity is $O(n_1n_2)$. The detailed discussion on the complexity and scalability of our SeedGNN is deferred to Appendix~\ref{sec:tscomplexity}.

\section{Theoretical Comparison Study}\label{sec:analysis}

We note that NGM (a supervised seedless GNN method) in \cite{wang2021neural}  bears some similarity with SeedGNN, because NGM also applies a pair-wise GNN, which uses an aggregation step similar to \prettyref{eq:kron-progagation}. However, a crucial difference is that after \prettyref{eq:kron-progagation}, NGM divides each node-pair representation by the product of the degrees of the corresponding two nodes. 
This type of normalization is quite common in GNNs to transform the non-topological features to a similar scale. However, this division can lead to very poor performance for seeded graph matching. 
The reason is that, if the number of seeds is small (\eg, $O(\log{n})$) and the node degree increases proportionally to $n$, after the normalization step in NGM, we expect that the resulting output value ($O(\frac{\log{n}}{n^2})$) will decrease to zero as $n$ increases. Hence, we expect that it would be difficult for NGM to distinguish the true pairs from the fake pairs in test graphs with larger sizes and node degrees than the training graphs. 

To formally study this effect, we conduct a theoretical study on a widely-adopted graph matching model, the correlated \ER graph model $\mathcal{G}(n,p,s;\theta)$ \cite{pedarsani2011privacy}. We first generate the parent graph $\mathcal{G}_0$ from the \ER model $\mathcal{G}(n, p)$ with $n$ nodes and edge probability $p$. Then, we obtain a subgraph $\mathcal{G}_1$ by sampling each edge of $\mathcal{G}_0$ independently with probability $s$.
Repeat the same sampling process independently to obtain another subgraph $\mathcal{G}_2$. 
Then, each true pair among $\mathcal{G}_1$ and $\mathcal{G}_2$ is independently added into the seed set $\mathcal{S}$ with probability $\theta$. We assume that the training set and test set have the same parameters $p,s,\theta$, and the only difference is the graph size, denoted by $\ntrain$ and $\ntest$, respectively. 

For ease of analysis, we focus on our SeedGNN model and an NGM-like model. For SeedGNN, we fix the number of layers $L=2$. The first layer is the same as we described in \prettyref{sec:method}. The second layer only uses the output of the percolation module as input (\ie, use $s_2=z_1$ instead of $s_2=[m_1,z_1]$). The NGM-like model is the same as SeedGNN except that the NGM-like model adds normalization after the aggregation step \prettyref{eq:kron-progagation}. Since the node degrees of \ER graphs are highly concentrated around the average degree, we let the NGM-like model divide all the node-pair representations by the square of the average degree. 
We assume that, after training, the Lipschitz constants of the neural networks $\rho_l$ (in \prettyref{eq:NN-rhol}) in NGM are all $K_L$.

We then present the sufficient conditions for the trained SeedGNN and NGM-like model to successfully match all nodes in test graphs.
Note that, in the first layer, both SeedGNN and the NGM-like model count the 1-hop witnesses in the same way, except that the NGM-like model further divides the output by the square of the average degree. Thus, the results are proportional to each other, and applying the Hungarian algorithm to these results yields the same matching results $R_1$. Let $\beta$ denote the fraction of correct matches in $R_1$ (which typically increases with the fraction of seeds $\theta$). 
However, the confidence levels  $Z_1$, computed by these two models are quite different because NGM does a normalization step. The difference in confidence levels will influence the performance in the second layer. More precisely, we have the following theorem, with the proofs deferred to
Appendix \ref{app:discuss-ngm}.

\begin{theorem}\label{thm:main}
Suppose $(\ntest ps)^2\geq c_0K_L\log{\ntest}$. We have
\begin{itemize}[noitemsep,topsep=0pt,leftmargin=5.5mm]
    \item if $\beta\geq c_1 \frac{\log \ntest}{\ntest ps^2}$, SeedGNN correctly matches all nodes with high probability; 
    \item if $\beta\geq c_2\max\left\{ \frac{\log{\ntest}}{\ntest ps^2},\ \sqrt{\frac{\log{\ntest}}{\ntest s^2}} \right\} $, the NGM-like model correctly matches all nodes with high probability,
\end{itemize}
where $c_0,c_1,c_2>1$ are some absolute constants.
\end{theorem}

Comparing the two sufficient conditions in \prettyref{thm:main}, we see that, when the graph is dense, i.e., $p \ge \sqrt{\frac{\log{\ntest}}{\ntest s^2}}$,  we have $\sqrt{\frac{\log{\ntest}}{\ntest s^2}}\geq \frac{\log \ntest}{\ntest ps^2}$ so that $\sqrt{\frac{\log{\ntest}}{\ntest s^2}} $ dominates the sufficient condition for the NGM-like model. Therefore, when the test graphs have large sizes and node degrees, the condition of SeedGNN require a much smaller $\beta$ (and thus fewer seeds) than that of NGM. Note that a smaller requirement of $\beta$ translates to a smaller number of initial seeds needed. Thus, given a small set of seeds, SeedGNN may already successfully match all nodes, while NGM still has a low matching accuracy (see \prettyref{sec:exp-res}).

\section{Experiments}\label{sec:experiment}

\subsection{Experimental Set-up}\label{sec:exp-set}

In our experiment, the number of SeedGNN layers is  fixed to 6. 
We implement the operators $\phi_l$ and $\rho_l$ as two-layer neural networks with $d_l= 16$. 
 For all experiments, optimization is done via ADAM \cite{kingma2014adam} with a fixed learning rate of $10^{-2}$. Our model is implemented using PyTorch \cite{NEURIPS2019_9015} and trained on an Intel Core i7-8750H CPU.  The performance is evaluated using the matching accuracy rate, \ie, the fraction of nodes that are correctly matched. Our code is publicly available at \url{https://github.com/Leron33/SeedGNN}. 
 
{\bf Datasets.}  We use the correlated \ER graph model \cite{pedarsani2011privacy}, Facebook networks in \cite{Traud_2012}, SHREC'16 computer vision dataset in \cite{lahner2016shrec}, and Willow Object dataset \cite{WillowObject} in our experiments. We described the correlated \ER graph model earlier in \prettyref{sec:analysis}, and the details of the three real datasets are deferred to \prettyref{app:dataset}.  

 {\bf Training set.} 
We construct the training set $\mathcal{T}$ in the following way. First, we generate 100 random pairs of correlated \ER graphs with $n= 100$, $p\in\{0.1, 0.3, 0.5\}$, $s\in\{0.6,0.8,1\}$, and $\theta=0.1$.
Second, we add 10 pairs of Facebook networks as discussed above with $\theta=0.1$ into the training set.
Third, we do not include any SHREC'16 dataset or Willow Object dataset in the training set. Our SeedGNN trained on the above training set already performs well for these two datasets (see \prettyref{sec:exp-res}), which verifies the generalization power of our SeedGNN. The training batch size is 64. The overall training for 500 epochs takes about 12 hours and requires 2.68 GB memory.

{\bf Baselines.} We compare the performance of our proposed SeedGNN with several state-of-the-art algorithms:  $\bm{D}$-hop \cite{mossel2019seeded}, PGM \cite{kazemi2015growing}, and PLD \cite{yu2021power} are theoretical algorithms;  SGM \cite{fishkind2019seeded} is  a convex relaxation algorithm;  MGCN \cite{chen2020multi} is a representative semi-supervised learning-based GNN approach; NGM \cite{wang2021neural} is a supervised GNN method for seedless graph matching. We adapt the NGM approach to seeded graph matching by replacing the affinity matrix in NGM with the Kronecker product of the two adjacent matrices and inputting the seed information in the same way as SeedGNN. The details of these baselines are deferred to \prettyref{app:baseline}.

\subsection{Results}\label{sec:exp-res}

\begin{figure*}[ht]
\vspace{-6pt}
\centering
\begin{subfigure}{0.4\textwidth}
\centering
 \includegraphics[width=0.9\columnwidth]{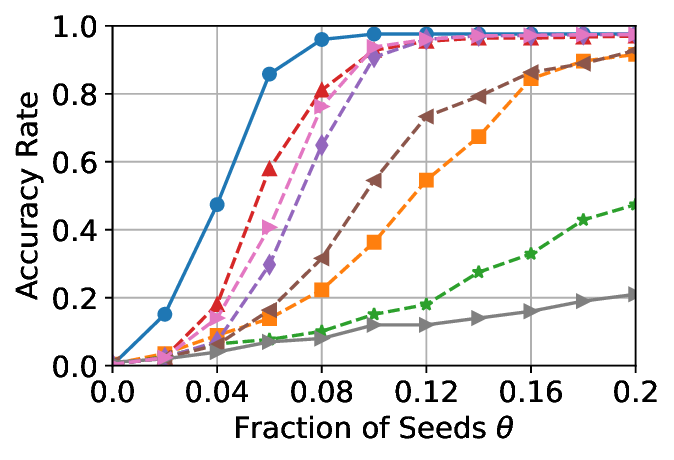}
 \caption{$p=0.01$}
 \label{fig:CompareER1}
\end{subfigure}
\quad
\begin{subfigure}{0.4\textwidth}
\centering
\includegraphics[width=0.9\columnwidth]{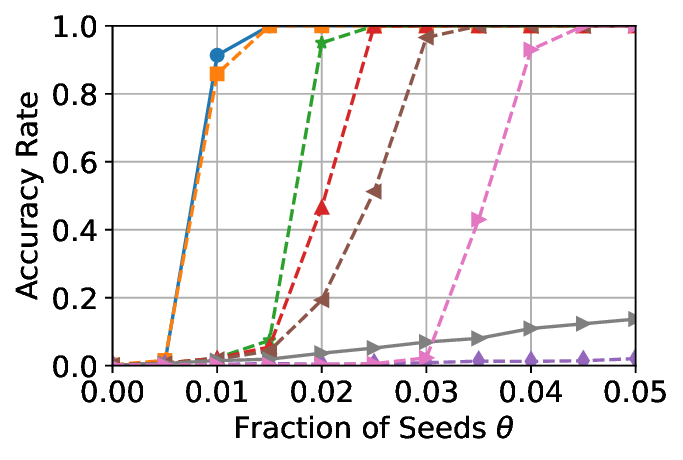}
\caption{$p=0.2$}
\label{fig:CompareER2}
\end{subfigure}
\raisebox{30pt}{
\includegraphics[width=0.15\columnwidth]{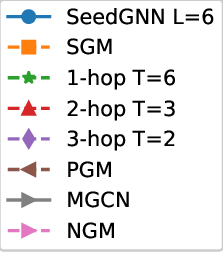}}
\caption{Performance comparison on correlated \ER graphs. Fix $n=500$ and $s=0.8$. 
}
\label{fig:CompareER}
\end{figure*}

\paragraph{SeedGNN requires fewer seeds than existing algorithms to successfully match graphs.}
In \prettyref{fig:CompareER}, we show the performance of the algorithms on the correlated \ER graph model. For test graphs, we vary $\theta$ while fixing $n=500$, $p\in\{0.01,0.2\}$, $s=0.8$. We can observe that, 
among the state-of-the-art methods, the iterative 2-hop algorithm has the best performance for sparse graphs ($p=0.01$), and the SGM algorithm performs the best for dense graphs ($p=0.2$). In comparison,
our SeedGNN has overall the best performance among all algorithms. 
Existing theoretical studies such as \cite{mossel2019seeded} must use witnesses at different numbers of hops, depending on whether the graphs are sparse ($p=0.01$) or dense ($p=0.2$). In contrast, our SeedGNN is capable of choosing the right features automatically to match different types of graphs. 

\begin{figure}[ht]
\centering
\includegraphics[width=0.75\columnwidth]{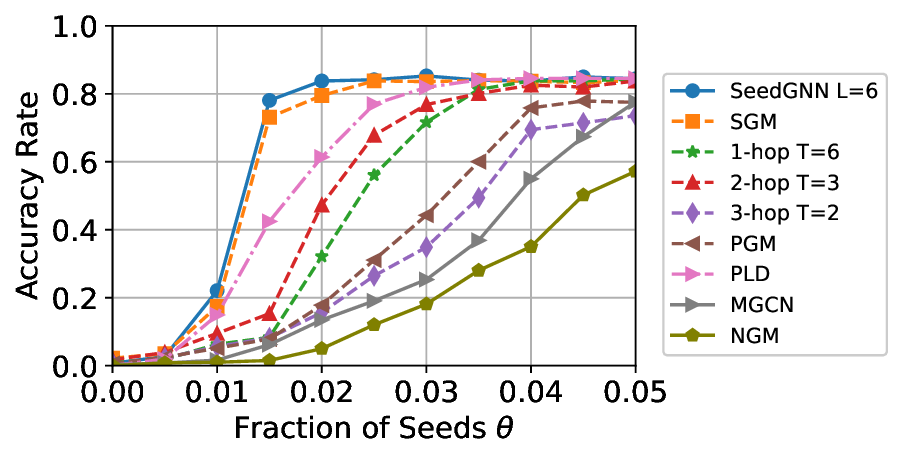}
\caption{Performance comparison  on the Facebook networks with different $\theta$.}
\label{fig:CompareFacebook}
\end{figure}

We then compare SeedGNN with the state-of-the-art algorithms on Facebook networks, which are real-world graphs with an approximate power-law degree distribution. In \prettyref{fig:CompareFacebook}, we can see that SeedGNN is comparable to SGM and significantly outperforms other algorithms. Note that the matching accuracy is saturated at around $80\%$, because there are about $15\%$ nodes that do not have any common neighbour in $\mathcal{G}_1$ and $\mathcal{G}_2$,
and thus can not be correctly matched. 

\begin{figure}[ht]
\centering
\includegraphics[width=0.75\columnwidth]{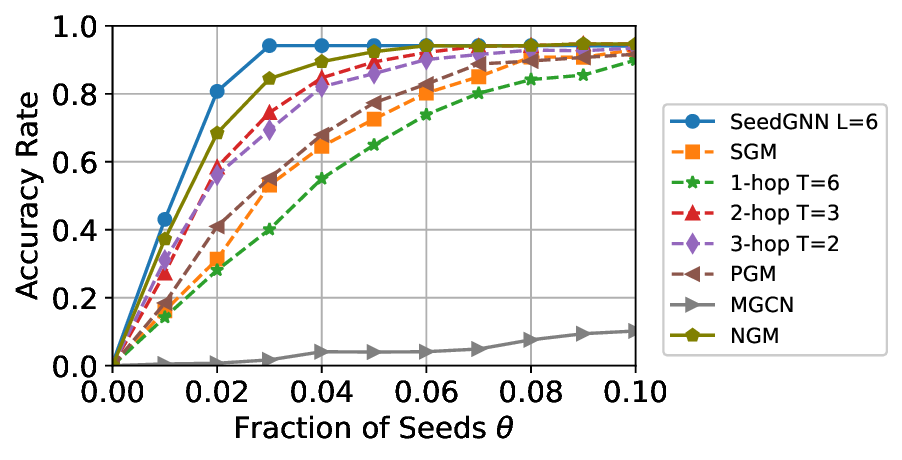}
\caption{Performance comparison on the SHREC'16 dataset with different $\theta$.}
\label{fig:CompareShrec}
\end{figure}

\paragraph{SeedGNN generalizes well to  unseen graphs with sizes and types different from the training graphs.} 
 When we test on the correlated \ER graphs in \prettyref{fig:CompareER}, the different graph sizes between the training set ($n=100$) and the test set ($n=500$) already demonstrates the generalization power of SeedGNN. To further validate that our SeedGNN can adapt to different graphs, we evaluate SeedGNN for deformable shape matching using the SHREC'16 dataset. Note that the sizes and types of graphs in this dataset are quite different from the \ER and Facebook graphs in the training set. The performance improvement shown in \prettyref{fig:CompareShrec} verifies the generalization power of our SeedGNN.

\paragraph{SeedGNN can be much more effective than semi-supervised learning GNN when non-topological features are not informative.} 
 Existing semi-supervised GNNs rely heavily on high-quality non-topological features (e.g., DeepLink \cite{zhou2018deeplink}, CrossMNA \cite{chu2019cross}, MGCN \cite{chen2020multi}). However, in the SHREC'16 dataset, the non-topological node features correspond to 3D coordinates, which do not provide much useful information for correlating two 3D shapes with different poses. As a result, in \prettyref{tab:CompareShrec}, we  observe that, provided with only a very small fraction of seeds ($\theta=0.01$), our SeedGNN can significantly outperform the semi-supervised methods.

\paragraph{Supervised node-based GNNs for seedless graph matching underperform for seeded graph matching.} 
We compare our SeedGNN with other supervised learning GNN methods on the SHREC'16 dataset.
We fix the fraction of seeds $\theta$ at 0.01, use the random encoding to represent seed information, and provide them as input for supervised GNNs (except NGM, for which we described the changes in \prettyref{sec:exp-set}). In \prettyref{tab:CompareShrec}, we can observe that our SeedGNN significantly outperforms the supervised GNNs, even when the latter are augmented with seed information.  This suggests that our method is more effective in using seed information than most supervised seedless GNN methods.

\paragraph{NGM does not generalize well when the test graph has much larger size and node-degree than the training graphs.}
As we discussed in \prettyref{sec:convolution}, most existing supervised seedless
GNN approaches are node-based  and have difficulty learning how to use seed information in a generalizable way. The only exception is NGM in \cite{wang2021neural}, which has a similar pair-wise architecture as SeedGNN. Thus, we transfer NGM to the seeded matching version. 
However, we can observe from \prettyref{fig:CompareER} that, although NGM performs relatively well (it still underperforms our SeedGNN) in larger sparse graphs ($p=0.01$), it performs quite poorly in larger dense graphs ($p=0.2$).  This observation confirms our theoretical analysis in \prettyref{sec:analysis}.
Note that, for the experiment on the SHREC'16 dataset (\prettyref{tab:CompareShrec}), NGM has similar performance as our SeedGNN. This is because in this experiment, we train NGM also with the SHREC '16 dataset. Thus, the test graphs and training graphs are with similar node degrees. As a result, the issue caused by the normalization operation in NGM is not as evident for the SHREC'16 dataset. 

\begin{table*}[ht]
\centering
\setlength\tabcolsep{2pt}
\caption{Comparison of GNN methods on SHREC’16 dataset.  The best results are marked as bold. }\label{tab:CompareShrec}
\scalebox{0.9}{
\begin{tabular}{|c|ccc|ccccc|}
\hline
\multirow{3}{*}{Method}& \multicolumn{3}{c|}{Semi-Supervised GNN}& \multicolumn{5}{c|}{Supervised GNN}\\
\cline{2-9}      & DeepLink & CrossMNA & MGCN & DGMC & BB-GM & DGM & NGM & SeedGNN \\[-3pt]
& \small \cite{zhou2018deeplink} & \small \cite{chu2019cross} & \small \cite{chen2020multi}  & \small\cite{fey2020deep}&\small\cite{rolinek2020deep}& \small\cite{gao2021deep} &\small \cite{wang2021neural} & ours \\ 
    \hline
{Accuracy (\%)}& $3.3 \pm 0.8$   & $4.2 \pm 1.7$ & $3.8 \pm 1.1$ &  $23.2\pm 6.8$ & $21.1\pm 4.4$ & $19.31\pm 10.6$  & $37.9\pm 5.7$  & $\bm{43.1\pm 8.5}$ \\\hline
\end{tabular}}\vspace{-6pt}
\end{table*}



\paragraph{SeedGNN can also be used in an overall pipeline to refine the outputs of other seedless graph matching algorithms,  taking advantage of informative non-topological features.}
Although SeedGNN is designed for seeded graph matching and only uses topological information, it can also be integrated into an overall pipeline to utilize non-topological node features that are sufficiently informative. For example, we can first use a seedless graph matching algorithm to generate an initial matching based on non-topological node features. 
By taking this initial matching as partially-correct seeds, we can then apply SeedGNN to correct these seeds.

To demonstrate this capability, we conduct experiments on the Willow Object dataset, whose node features are informative enough for correlating nodes. We compare the SeedGNN pipeline with several state-of-the-art seedless GNN methods (see \prettyref{tab:gnn}).
The performance values of these existing seedless GNN methods are directly retrieved from their respective papers. 
For the SeedGNN pipeline, we still directly use the model trained in \prettyref{sec:exp-set}. The input to SeedGNN is generated in two ways. The first way is to apply a neural network only on non-topological features (without graph information) and use the dot product of its outputs on each pair of nodes to generate the similarity for each node-pair (we denote this neural network as \textbf{MLP}). Then, \textbf{MLP+SeedGNN-1} directly uses the similarity values of all node-pairs as input, while \textbf{MLP+SeedGNN-2} further filters out noisy information by applying the Hungarian algorithm on the similarity matrix. The second way (\textbf{DGMC+SeedGNN}) is to use the matching result of DGMC, one of the GNN methods for seedless graph matching, as input. 

From the matching accuracy presented in \prettyref{tab:gnn}, we make the following observations.
First, our results show that SeedGNN can effectively refine the output of other seedless graph matching algorithms. Specifically, both MLP+SeedGNN-1 and MLP-SeedGNN-2 outperform MLP, while DGMC+SeedGNN outperforms DGMC. Second, it is crucial to clean up the output of seedless matching algorithms in order to achieve good performance in the SeedGNN pipeline. This is evident as MLP+SeedGNN-2 consistently outperforms MLP+SeedGNN-1. Indeed, the similarity matrix contains significantly more fake pairs than true pairs. Thus, by using the Hungarian algorithm on the similarity matrix, MLP+SeedGNN-2 sends only high-confidence seeds to SeedGNN, and is more effective in suppressing the misleading information. Third, our experiments show that the effectiveness of the SeedGNN pipeline still depends on the choice of seedless algorithm. Specifically, we observe that DGMC+SeedGNN achieves the best performance, outperforming MLP+SeedGNN and other seedless GNN methods.



\begin{table}[ht]
\vspace{-6pt}
\centering
\setlength\tabcolsep{1pt}
\caption{Comparison of matching accuracy (\%) on Willow Object dataset. The best results are marked as bold. 
}
\scalebox{1}{
\begin{tabular}{|c|ccccc|c|}
\hline
Method &  face & mbike & car & duck & wbottle & Mean \\ \hline 
GMN \small \cite{zanfir2018deep}& 98.1 & 65.0 & 72.9 & 74.3 & 70.5 & 76.2 \\ 
PCA-GM \small \cite{wang2019learning} & \textbf{100.0} & 76.7 & 84.0 & 93.5 & 96.9 & 90.2 \\ 
NGM \small \cite{wang2021neural} & 99.2  & 82.1  & 84.1  & 77.4  & 93.5  & 87.2 \\
IPCA-GM \small \cite{wang2020combinatorial}& \textbf{100.0} & 77.7 & 90.2 & 84.9 & 95.2 & 89.6\\
CIE \small \cite{yu2019learning} &  \textbf{100.0} & 90.0 & 82.2 & 81.2 & 97.6 & 90.2\\ 
DGMC \small  \cite{fey2020deep}& \textbf{100.0} & 92.1 & 90.3 & 89.0 & 97.1 & 93.7 \\ 
BB-GM \small  \cite{rolinek2020deep}   & \textbf{100.0} & 98.9 & 95.7 & 93.1 & 99.1 & 97.4 \\ 
DGM \small \cite{gao2021deep}   & \textbf{100.0} & 98.8 & \textbf{98.0} & 92.8 & 99.0 & 97.7\\
DLGM \small \cite{pmlr-v139-yu21d}  & \textbf{100.0}  & 99.3 & 96.5 & 93.7 & \textbf{99.3} & 97.8\\
MLP & 98.1 & 48.3 & 65.3 & 66.0 & 77.7& 71.1 \\\hline
 MLP+SeedGNN-1 & 99.4 & 77.8 & 84.1 & 77.4 & 89.5& 85.6\\
 MLP+SeedGNN-2 & \textbf{100.0} & 98.9 & 96.8 & 93.1 & 98.7 & 97.5\\
 DGMC+ SeedGNN & \textbf{100.0} & \textbf{99.6} & 97.4 & \textbf{98.7} & 99.0 & \textbf{98.9}\\
\hline
\end{tabular}}
\label{tab:gnn}
\vspace{-10pt}
\end{table}

\paragraph{Additional experiments to study the complexity and inner working of our SeedGNN.}  In \prettyref{app:scalable}, we show that the runtime of our SeedGNN is comparable to other GNN-based algorithms. In \prettyref{app:innerworking}, we investigate the inner working of our SeedGNN. First, we  verify the effectiveness of our design choices for SeedGNN by comparing the performance of different architectural designs. 
Then, we investigate which sets of samples need to be included in our training set to obtain an effective trained model. 
Finally, we study the matching process of SeedGNN for different types of graphs. The results verify that SeedGNN chooses the appropriate features for different graphs based on the confidence levels of new seeds as illustrated in~\prettyref{fig:layerwise-feature}.

In summary, all of the above experiments demonstrate that our SeedGNN significantly outperforms these baselines across various types of graphs 
while requiring fewer seeds.
Moreover, the knowledge learned by SeedGNN from training graphs can be effectively generalized to test graphs of different sizes and categories.

Limitations and societal impacts are discussed in the Appendix \ref{app:limitations} and \ref{app:impact}, respectively.

\section*{Acknowledgements}
L.~Yu and X.~Lin are supported in part by the NSF Grants CNS-2113893 and CNS-2225950. J.~Xu is supported in part by the NSF Grant CCF-1856424 and an NSF CAREER award CCF-2144593.
We would like to thank Prof. Qiang Qiu at Purdue University and the anonymous reviewers for their valuable comments and suggestions on our paper.

\bibliography{main}
\newpage
\appendix

\section{Additional Discussion on Related Work}\label{app:additional-related-work}

\paragraph{Theoretical Algorithms } 
Existing theoretical algorithms suffer from several limitations.
First, graphs with different characteristics may require different features and carefully tuned parameters. For instance, the Percolation algorithm in \cite{kazemi2015growing} needs to carefully choose a threshold parameter to achieve good performance, and the PLD algorithm in \cite{yu2021power} is designed for graphs with power-law degree distributions.   It is cumbersome to design a new algorithm and/or tune the parameters every time when a new type of graph is processed. In contrast, by learning from the training graphs, our SeedGNN can automatically choose the effective features. Second, these theoretical algorithms may not synthesize different features most effectively.  For instance, the $l$-hop algorithm in \cite{mossel2019seeded} only utilizes the  witnesses at the $l$-th hop, but does not combine witnesses at different hops. 
 

\paragraph{Inductive Semi-supervised Learning on Graphs}
 Our goal of using supervised learning for seeded graph matching shares some similarities with the work in \cite{wen2021meta}, which also aims to both perform inductive learning (i.e., learn transferrable knowledge from training graphs) and utilize a small amount of labeled data on the test graph. However, \cite{wen2021meta} focuses on a node classification problem, which is quite different from seeded graph matching. In particular, \cite{wen2021meta} uses node-based GNNs, which (as we discussed in Section 4.1) have more difficulty in effectively utilizing seed information than our proposed pair-wise GNN. Further, in order to transfer knowledge from the trained GNN to test graphs, \cite{wen2021meta} scales all GNN weights by a common factor. It is unclear how this scaling will effectively transfer knowledge for seeded graph matching, e.g., how to best use different hops of witnesses. In contrast, our design of SeedGNN exploits the inherent structure of the seeded graph matching problem, and can be shown to generalize well to unseen graphs of sizes and types very different from the training set.  For future work, it would be of interest to explore whether our SeedGNN can be further improved with a meta-learning component  \cite{santoro2016meta}. 

\paragraph{Convex Relaxation Algorithms} In addition to the theoretical algorithms and the GNN approaches, there is another class of algorithms based on convex relaxations of the quadratic assignment problem, which maximizes the total number of matched edges between two graphs subject to the seed constraint \cite{lyzinski2014seeded,fishkind2019seeded}.
In \cite{fishkind2019seeded}, the authors describe a gradient ascent approach to solve this relaxed problem, which is called SGM. 
Compared to SeedGNN, SGM also has flavors of using witnesses and percolation ideas.
Specifically, the gradient of the SGM algorithm is similar to a matrix counting 1-hop witnesses. However, using only 1-hop witnesses is known to be ineffective in sparse graphs (as there are very few 1-hop witnesses even for true pairs).
Indeed, our experiments in \prettyref{sec:experiment} find that our SeedGNN often outperforms SGM, especially in sparse graphs.


\paragraph{Differences between Our Percolation Module and Previous Percolation Algorithms} Unlike previous percolation algorithms \cite{yartseva2013performance}, we allow SeedGNN to correct errors from earlier layers by re-matching nodes at each layer. Note that in many percolation algorithms, once a new pair of seeds is identified, it will be used as the correct match until the end. This approach can be problematic if an incorrect pair is identified as a seed, whose impact will be lasting for many iterations down the road. In contrast, since our SeedGNN rematches nodes at each layer, even if some of the newly-identified seeds in the previous layer are incorrect, we can correct these errors in the next layer, as long as the fraction of incorrect seeds is small. In other words, our design of SeedGNN takes advantage of the power of partially correct (i.e., noisy) seeds (as theoretically verified in \cite{lubars2018correcting,yu2021graph}).

\section{Proof of \prettyref{thm:main}}\label{app:discuss-ngm}

\paragraph{Notation}
We use $\overset{\cdot}{\sim}$ to denote ``approximately distributed".

Note that the first layers of SeedGNN and the NGM-like model output the same set of new seeds ($\beta$ fraction of which are correct). However, their confidence levels are very different. Specifically, after softmax in \prettyref{eq:normalize-softmax}, correctly matched pairs will have confidence levels close to $1$, while incorrectly matched pairs will only have confidence levels close to $\frac{1}{\ntest}$. In contrast, since the NGM-like model divides the node-pair representations by the square of the average degree, the resulting value ($\sim\frac{1}{\ntest^2}$) will decrease close to zero as $\ntest$ increases. After we apply softmax, the confidence levels of correct matching and incorrect matching will both become close to $\frac{1}{\ntest}$. More precisely, we have the following lemma, with the proof deferred to \prettyref{app:proofs}.
\begin{lemma}\label{lmm:NGM-softmax}
Assume that $(\ntest ps)^2\geq c_0K_L\log{\ntest}$ for a sufficiently large constant $c_0$. 
In the first layer of the NGM-like model, for any $(i,j)$, we have $\frac{1-\delta}{\ntest}\leq {Y}_1(i,j)\leq \frac{1+\delta}{\ntest}$ with high probability, where $\delta$ is some sufficiently small constant.
\end{lemma}

Since the NGM-like model has lower confidence levels for the correctly-matched new seeds than SeedGNN, it is more difficult for the NGM-like model to match all nodes correctly in the second layer. We present the sufficient conditions for the second layer of SeedGNN and NGM-like model to correctly match all nodes. The proofs of these theorems are deferred to \prettyref{app:proofs}.

\begin{theorem}\label{thm:seedgnn}
If $\beta \geq c_1 \frac{\log \ntest}{\ntest ps^2}$, SeedGNN correctly matches all nodes with high probability.
\end{theorem}

\begin{theorem}\label{thm:NGM}
 If $(\ntest ps)^2\geq c_0K_L\log{\ntest}$ and $\beta \geq c_2 \max\left\{ \frac{\log{\ntest}}{\ntest ps^2},\ \sqrt{\frac{\log{\ntest}}{\ntest s^2}} \right\}$, the NGM-like model correctly matches all nodes with high probability.
\end{theorem}

\subsection{Postponed Proofs}\label{app:proofs}

\begin{proof}[Proofs of \prettyref{lmm:NGM-softmax}]
Recall that the first layer of the NGM-like model counts the 1-hop witnesses and divides the value by the square of average degree. Thus, in testing, we have, for any node-pair $(i,j)$,
$$
 H_1(i,j)=\frac{1}{(\ntest ps)^2}\cdot\sum_{(u,v)\in \mathcal{S}}A_1(i,u)A_2(v,j).
 $$

For any initial seed $(u,v)$, $A_1(i,u)A_2(v,j)$ is equal to 1 with probability $ps^2$ if $j=\pi(i)$, and $A_1(i,u)A_2(v,j)$ is equal to 1 with probability $p^2s^2$ if $j\neq \pi(i)$. Since there are $\theta$ fraction of seeds, $ H_1(i,j)$ follows the distribution given by
\begin{subequations}
   \begin{empheq}[left={H_1(i,j)\overset{\cdot}{\sim}\empheqlbrace\,}]{align}
        &\frac{1}{(\ntest ps)^2}\Binom\left(\ntest\theta,ps^2\right)& \text{if } j&=\pi(i),   \nonumber\\
    &\frac{1}{(\ntest ps)^2}\Binom(\ntest\theta,p^2s^2)& \text{if }  j&\neq \pi(i).\nonumber
  \end{empheq} 
\end{subequations}

By Bernstein's Inequality \cite{Dubhashi2009ConcentrationOM}, the upper bound of $H_1(i,j)$ is 
\begin{align*}
&\prob{H_1(i,j)>\frac{\ntest\theta ps^2+\sqrt{6\ntest\theta ps^2\log{\ntest}}+2\log{\ntest}}{(\ntest ps)^2}}\\
\leq &\prob{\Binom\left(\ntest\theta,ps^2\right)>\ntest\theta ps^2+\sqrt{6\ntest\theta ps^2\log{\ntest}}+2\log{\ntest}}\\
\leq &\exp\left(-3\log{\ntest}\right) <\ntest^{-3}.
\end{align*}

Recall that we apply a neural network $\rho_1$ on each element of $H_1$ to get $X_1=\rho_1(H_1)$, and the Lipschitz constant of $\rho_1$ is $K_L$. Since $H_1(i,j)\geq 0$ for any $(i,j)$, we have 
$$\max_{(i,j),(u,v)}\abs{X_1(i,j) - X_1(u,v)}\leq K_L \max_{(i,j),(u,v)}\abs{H_1(i,j) - H_1(u,v)}\leq K_L \max_{(i,j)}H_1(i,j).
$$
Thus, we let $\frac{\delta}{2}=K_L \frac{\ntest\theta ps^2+\sqrt{6\ntest\theta ps^2\log{\ntest}}+2\log{\ntest}}{(\ntest ps)^2}$ and have
\begin{align}\label{eq:X-bound}
\prob{\max_{(i,j),(u,v)}\abs{X_1(i,j) - X_1(u,v)}> \frac{\delta}{2}}
\leq & \prob{ K_L \max_{(i,j)}H_1(i,j)> \frac{\delta}{2}}\nonumber\\
\leq&\prob{\bigcup_{(i,j)}\{K_L H_1(i,j)> \frac{\delta}{2}\}}\nonumber\\
\leq &{\ntest}^2\prob{ H_1(i,j)> \frac{\ntest\theta ps^2+\sqrt{6\ntest\theta ps^2\log{\ntest}}+2\log{\ntest}}{(\ntest ps)^2}}\nonumber\\
\leq& \ntest^{-1}.
\end{align}
Since $\ntest p\geq \theta$ and $(\ntest ps)^2\geq c_0K_L\log{\ntest}$, $\frac{\delta}{2}$ can be made to be sufficiently small and $1\leq\exp(\frac{\delta}{2})<1+\delta$. It follows taht
$$
\frac{\exp(\frac{\delta}{2})}{\exp(\frac{\delta}{2})+(\ntest-1)\exp(0)} \leq \frac{1+\delta}{\ntest}\quad \text{and}\quad \frac{\exp(0)}{\exp(\frac{\delta}{2})+(\ntest-1)\exp(0)}\geq  \frac{1}{\ntest+\delta}\geq \frac{1-\delta}{\ntest}.
$$

We then apply row-wise softmax on $X_1$ to get the confidence level $Y_1$. We can bounded $Y_1$ by the difference between elements in $X_1$:
\begin{align*}
\max_{(i,j)} Y_1(i,j)\leq& \frac{\exp(\max_{(i,j)}X_1(i,j))}{\exp(\max_{(i,j)}X_1(i,j))+(\ntest-1)\exp(\min_{(i,j)}X_1(i,j))}\\
=&\frac{\exp(\max_{(i,j)}X_1(i,j)-\min_{(i,j)}X_1(i,j))}{\exp(\max_{(i,j)}X_1(i,j)-\min_{(i,j)}X_1(i,j))+(\ntest-1)\exp(0)},
\end{align*}
and similarly
\begin{align*}
    \min_{(i,j)} Y_1(i,j)\geq \frac{\exp(0)}{\exp(\max_{(i,j)}X_1(i,j)-\min_{(i,j)}X_1(i,j))+(\ntest-1)\exp(0)}.
\end{align*}
Since the difference in $X_1$ is upper bounded in \prettyref{eq:X-bound}, we have 
\begin{align*}
    \prob{\max_{(i,j)} Y_1(i,j) \leq \frac{1+\delta}{\ntest}}\geq&\prob{\max_{(i,j)} Y_1(i,j)\leq \frac{\exp(\frac{\delta}{2})}{\exp(\frac{\delta}{2})+(\ntest-1)\exp(0)} }\\
    \geq &1-\prob{\max_{(i,j),(u,v)}\abs{X_1(i,j) - X_1(u,v)}> \frac{\delta}{2}} \\ \geq& 1-\ntest^{-1},
\end{align*}
\begin{align*}
     \prob{\min_{(i,j)} Y_1(i,j) \geq \frac{1-\delta}{\ntest}}
     \geq&\prob{\min_{(i,j)} Y_1(i,j)\geq\frac{\exp( 0)}{\exp(\frac{\delta}{2})+(\ntest-1)\exp(0)}}\\
      \geq &1-\prob{\max_{(i,j),(u,v)}\abs{X_1(i,j) - X_1(u,v)}> \frac{\delta}{2}}\\
     \geq &1-\ntest^{-1}.
\end{align*}
    
\end{proof}

\begin{proof}[Proof of \prettyref{thm:seedgnn}]
 Recall that the second layer of SeedGNN aggregates over node-pair's neighborhoods in \prettyref{eq:matrix-mul}. Then, we have, for any node-pair $(i,j)$,
 $$
 H_2(i,j)=\sum_{(u,v):u,v\in [\ntest]}A_1(i,u)A_2(v,j)S_2(u,v).
 $$

In our analysis, we use the output of percolation module in the first layer as input (\ie $S_2=Z_1$).
Note that $Z_1$ is the ``cleaned'' confidence levels. There are only $n_1$ non-zeros elements in $Z_1$ representing the confidence levels of new seeds. Among $n_1$ new seeds, $\beta$ fraction of them are correctly matched seeds, and the rest are incorrectly matched seeds. Since correct seeds have much higher confidence levels than incorrect seeds, after softmax normalization in \prettyref{eq:normalize-softmax}, correct seeds have 1 confidence level and incorrect seeds have $\frac{1}{\ntest}$ confidence level. Then,
 We have
$$
H_2(i,j)=\sum_{\substack{(u,v):Z_1(u,v)>0,\\ v=\pi(u)}}A_1(i,u)A_2(v,j)+\frac{1}{\ntest}\cdot\sum_{\substack{(u,v):Z_1(u,v)>0,\\ v\neq\pi(u)}}A_1(i,u)A_2(v,j).
 $$
For any correct seed, $A_1(i,u)A_2(v,j)$ is equal to 1 with probability $ps^2$ if $j=\pi(i)$, and $A_1(i,u)A_2(v,j)$ is equal to 1 with probability $p^2s^2$ if $j\neq \pi(i)$. In contrast, for any incorrect seed, $A_1(i,u)A_2(v,j)$ is equal to 1 with probability $p^2s^2$ for any node-pair $(i,j)$. Thus, it follows that 
\begin{subequations}
   \begin{empheq}[left={H_2(i,j)\overset{\cdot}{\sim}\empheqlbrace\,}]{align}
        &\Binom\left(\ntest\beta,ps^2\right)+ \frac{1}{\ntest}\Binom\left(\ntest(1-\beta),p^2s^2\right)& \text{if } j&=\pi(i),   \nonumber\\
    &\Binom(\ntest\beta,p^2s^2)+\frac{1}{\ntest}\Binom\left(\ntest(1-\beta),p^2s^2\right) & \text{if }  j&\neq \pi(i).\nonumber
  \end{empheq} 
\end{subequations}
Since the second terms are both no greater than 1,  $H_2(i,j)$ is dominated by the first terms in the right-hand-side (which are contributed by the correct seeds). If  $H_2(i,j)$ of any true pair is greater than  $H_2(i,j)$ of any fake pair, all the true pairs can be distinguished from fake pairs.  By Theorem 1 in \cite{korula2014efficient}, if $\beta\geq c_1\frac{\log \ntest}{\ntest ps^2}$ with a sufficiently large constant $c_1$, all nodes can be correctly matched with high probability.

\end{proof}

\begin{proof}[Proof of \prettyref{thm:NGM}]
In the NGM-like model, the second layer aggregates over node-pair's neighborhoods and divides the results by the square of average degree. Thus, we have, for any node-pair $(i,j)$,
 $$
 H_2(i,j)=\frac{1}{(\ntest ps)^2}\sum_{(u,v):u\in [n_1],v\in [n_2]}A_1(i,u)A_2(v,j)S_2(u,v).
 $$
 
We use the ``cleaned" matching as input (\ie, $S_2=Z_1$), and there are only $n_1$ non-zeros elements in $Z_1$ representing the confidence levels of new seeds. Among the $n_1$ new seeds, $\beta$ fraction of them are correctly matched seeds, and the rest are incorrectly matched seeds. By \prettyref{lmm:NGM-softmax}, the confidence levels of all new seeds are close to $\frac{1}{\ntest}$ with high probability. Then,
 We have
$$
H_2(i,j)=\frac{1}{\ntest(\ntest ps)^2}\cdot\left(\sum_{\substack{(u,v):Z_1(u,v)>0,\\ v=\pi(u)}}A_1(i,u)A_2(v,j)+\sum_{\substack{(u,v):Z_1(u,v)>0,\\ v\neq\pi(u)}}A_1(i,u)A_2(v,j)\right).
 $$

Analogous to the analysis in the proof of \prettyref{thm:seedgnn},
 $H_2(i,j)$ follows the distribution given by
\begin{subequations}
   \begin{empheq}[left={H_2(i,j)\overset{\cdot}{\sim}\empheqlbrace\,}]{align}
        &\frac{1}{\ntest(\ntest ps)^2}\left(\Binom\left(\ntest\beta,ps^2\right)+ \Binom\left(\ntest(1-\beta),p^2s^2\right)\right)& \text{if } j&=\pi(i),   \nonumber\\
    &\frac{1}{\ntest(\ntest ps)^2}\left(\Binom(\ntest\beta,p^2s^2)+\Binom\left(\ntest(1-\beta),p^2s^2\right)\right)  & \text{if }  j&\neq \pi(i).\nonumber
  \end{empheq} 
\end{subequations}   
If  $H_2(i,j)$ of any true pair is greater than  $H_2(i,j)$ of any fake pair, all the true pairs can be distinguished from fake pairs by the NGM-like model.
Note that we need to consider the influence of incorrect seeds. By Theorem 1 in \cite{yu2021graph}, 
if $\beta\geq c_2 \max\left\{ \frac{\log{\ntest}}{\ntest ps^2},\ \sqrt{\frac{\log{\ntest}}{\ntest s^2}} \right\}$ with a sufficiently large constant $c_2$, all nodes can be correctly matched with high probability.
\end{proof}

\section{Complexity and Scalability}\label{sec:tscomplexity}

\subsection{Time and Space Complexity}

First, we analyze the computational complexity of our SeedGNN. 
In each layer, $(A_1\otimes A_2)s_l$ in \prettyref{eq:kron-progagation} can be converted into $A_1S_lA_2$ as shown in \prettyref{eq:matrix-mul}. When we represent $A_1$ and $A_2$ with sparse matrices, each of them only contain $n_1d_{\text{mean}}$ and $n_2d_{\text{mean}}$ elements, where $d_{\text{mean}}$ is the mean of the node degrees of $\mathcal{G}_1$ and $\mathcal{G}_2$. Thus, by sparse matrix multiplication, the time complexity of Equation (3) is $O(n_1n_2d_{\text{mean}})$.
The neural networks \prettyref{eq:NN-hl} and \prettyref{eq:NN-rhol} take $O(n_1n_2)$ time. The Hungarian algorithm takes $O(n_1n_2^2)$ times \cite{crouse2016implementing}. Thus, the total time complexity is $O(n_1n_2^2)$.

The space complexity of our SeedGNN is $O(n_1n_2)$, since $A_1$ and $A_2$ are sparse matrices, and $S_l$ has $n_1n_2$ elements.

\subsection{Making SeedGNN more Scalable}\label{app:scalable}

For very large graphs, the step of the Hungarian algorithm  becomes the computational bottleneck. We can use greedy max-weight matching  (GMWM) in \cite{avis1983survey} instead, as the time complexity of GMWM is only $O(n_1n_2 \log n_2)$. With this improvement, the total time complexity is reduced to $O(n_1n_2 \log n_2+n_1n_2d_{\mean})$.
To the best of our knowledge, the best-known time complexity for GNN-based algorithms is $O(n_1n_2)$ \cite{fey2020deep}.
Thus, the computational complexity of our SeedGNN is only moderately larger than the best-known one. In \prettyref{tab:runtime}, we show the average run time of GNN-based algorithm to match a pair of large graphs on SHREC’16 dataset (with 8K-11K nodes). The semi-supervised methods (DeepLink \cite{zhou2018deeplink}, CrossMNA \cite{chu2019cross}, MGCN \cite{chen2020multi}) are provided with a fraction of seeds ($\theta=0.01$) as the training set, and the run time includes the training and test time. The supervised methods (DGMC \cite{fey2020deep}, BB-GM \cite{rolinek2020deep}, DGM \cite{gao2021deep}, NGM \cite{wang2021neural}) utilize only non-topological features but not seeds, and the run time is only is only for a pair of test graphs. The results  demonstrate that the run time of our SeedGNN is comparable to the best-known GNN-based algorithms.

\begin{table}[ht]
\vspace{-5pt}
\centering
\setlength\tabcolsep{1.5pt}
\caption{Run time Comparison of GNN methods on SHREC’16 dataset.}\label{tab:runtime}
\vspace{-3pt}
\scalebox{0.85}{
\begin{tabular}{|c|ccc|ccccc|}
\hline
\multirow{3}{*}{Method}& \multicolumn{3}{c|}{Semi-Supervised}& \multicolumn{5}{c|}{Supervised}\\
\cline{2-9}  & DeepLink & CrossMNA & MGCN & DGMC & BB-GM & DGM & NGM & SeedGNN \\[-4pt]    & \small \cite{zhou2018deeplink} & \small \cite{chu2019cross} & \small \cite{chen2020multi}  & \small\cite{fey2020deep} & \small \cite{rolinek2020deep}& \small \cite{gao2021deep}& \small \cite{wang2021neural} & \small ours \\ \hline run time (s)& $1847.0$   & $2321.9$ & $3573.4$ & $80.2$ & $130.1$ & $211.3$ & $879.2$  & $141.5$ \\\hline
\end{tabular}}
\end{table}

\section{Details of Experiments}

\subsection{Datasets}\label{app:dataset}
We give detailed descriptions of the real datasets used in our experiments.
\paragraph{Facebook  networks} The dataset in \cite{Traud_2012} provides 100 Facebook networks from different institutions. We randomly choose 10 for training and 90 for testing. The sizes of the Facebook networks range from 962 to 32361. To lower the training cost, we down-sample the sizes of the training graphs. Specifically, for each Facebook network for training, we first down-sample nodes with probability $0.25$ to get the parent graph $\mathcal{G}_0$. However, for testing, we do not perform this down-sampling and use the original graphs directly as the parent graph $\mathcal{G}_0$. For both training and testing, we generate $\mathcal{G}_1$ and $\mathcal{G}_2$ from $\mathcal{G}_0$ by independently sub-sampling each edge of $\mathcal{G}_0$ twice with probability $s=0.8$ and sub-sampling each node of $\mathcal{G}_0$ twice with probability $0.9$.
The nodes of $\mathcal{G}_2$ are then relabeled according to a random permutation $\pi$. Then, each true pair is independently added into the seed set $\mathcal{S}$ with probability $\theta$.

\paragraph{
The SHREC'16 dataset} Matching 3D deformable shapes is a central problem in computer vision, and has been extensively studied for decades (see~\cite{van2011survey} and~\cite{sahilliouglu2020recent} for
surveys). The SHREC'16 dataset in \cite{lahner2016shrec} provides 25 deformable 3D shapes (15 for training and 10 for testing) undergoing different topological changes. Each shape is represented by a triangulated mesh graph consisting of around 8K-11K nodes (with 3D coordinates).

\paragraph{Willow Object dataset } Willow Object dataset \cite{WillowObject} contains
at least 40 images for each of its five categories.  Following the experimental setups in \cite{fey2020deep}, we construct graphs via the Delaunay triangulation of keypoints, and each image consists of exactly 10 labeled keypoints. The features of the keypoints are given by the concatenated output of relu4\_2 and relu5\_1 of a pre-trained VGG16 \cite{Simonyan2015VeryDC}.

\subsection{Details of Baselines}\label{app:baseline}
\textbf{1) $\bm{D}$-hop} \cite{mossel2019seeded} finds the node mapping between  the  two  graphs  that  maximizes  the  total  number  of $D$-hop witnesses for a given $D$. For a fair comparison with other algorithms, we iteratively apply the $D$-hop algorithm $T$ times (with $DT=6$ because SeedGNN is fixed to have 6 layers). In each iteration, we use the matching result of the previous iteration as new seeds and apply the $D$-hop algorithm again.   \textbf{2) PGM} \cite{kazemi2015growing}  iteratively matches node-pairs with at least $r$ witnesses. We choose $r=2$, which is the same as the simulation setting in \cite{kazemi2015growing}. \textbf{3) PLD} \cite{yu2021power} is the state-of-the-art seeded graph matching algorithm designed for graphs with power-law degree distributions (which is a common feature of real-world social networks \cite{barabasi2016network}).  \textbf{4) SGM} \cite{fishkind2019seeded} uses Frank–Wolfe method to approximately solves a quadratic assignment problem that maximizes the number of matched edges between two graphs, while being consistent with the given seeds.  \textbf{5) MGCN} \cite{chen2020multi} is a representative semi-supervised learning-based GNN approach, whose performance is comparable with other semi-supervised learning approaches. The parameters are set in the same way as those in \cite{chen2020multi}. \textbf{6) NGM} \cite{wang2021neural} is a supervised GNN method for seedless graph matching, but it also uses a  pair-wise GNN that utilizes an affinity matrix as input. We transfer this approach to seeded graph matching by replacing the affinity matrix in NGM by the Kronecker product of the two adjacent matrices, and inputting the seed information as SeedGNN. We then train the weights of NGM with the same training set as our SeedGNN. 

\section{Studying the Inner-working of SeedGNN}\label{app:innerworking}

In this section, we further investigate how the performance of SeedGNN varies as we change its inner working. First, to verify the effectiveness of our design choices for our SeedGNN method, we compare the performance of different architectural designs. Then, we investigate which sets of samples need to be included in our training set to obtain an effective trained model. 
Finally, we study the matching process of SeedGNN for different types of graphs. The results suggest that SeedGNN could potentially choose the appropriate features for different graphs based on the confidence levels of new seeds.

\subsection{Study of the Design Choices}\label{app:exp-design}

To verify the effectiveness of our design choices, we consider four variants of SeedGNN, which are: 
\begin{enumerate}
    \item {\bf SeedGNN-x:} SeedGNN without convolution module. This variant aims to verify the importance of extracting witness-like information at a larger number of hops.
    \item {\bf SeedGNN-w:} SeedGNN without percolation module. This variant aims to verify the importance of the percolation module in SeedGNN. 
    \item {\bf SeedGNN-p:} SeedGNN with percolation module but without the Hungarian matching algorithm (\ie, $z_l=\mathsf{unflatten}(Y_l)$ in each GNN layer).  This variant aims to verify the importance of the ``cleaning" process in SeedGNN. 
    \item {\bf SeedGNN-h:} SeedGNN with $z_l=\mathsf{unflatten}(R_l)$ instead of \prettyref{eq:mask} in each layer. This variant aims to verify that among the new seeds, it is still important to distinguish the high-confidence one and low-confidence one.
\end{enumerate}  
Finally, we use ``SeedGNN'' to denote the full design in Fig.~\ref{fig:Architecture}.   We train all these variants with the same training set $\mathcal{T}$  in \prettyref{sec:exp-set}. 

\begin{figure}[ht]
\centering
\includegraphics[width=0.7\columnwidth]{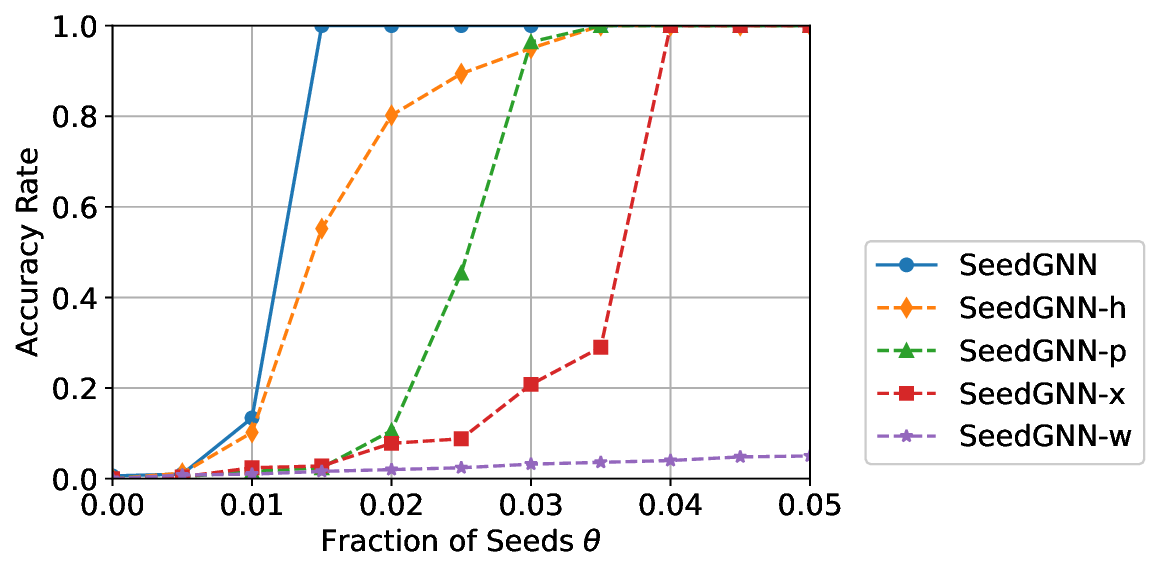}
\caption{Performance comparison of our SeedGNN and four other variants on correlated \ER graph model with different $\theta$. Fix $n=500$, $p=0.04$, $s=0.8$.}
\label{fig:CompareDesign}
\end{figure}

In \prettyref{fig:CompareDesign}, we show the performance of the above variants of SeedGNN on correlated \ER graph model. For test graphs,  we increase $\theta$ from $0$ to $0.05$ while fixing $n=500$, $p=0.04$, $s=0.8$.   As illustrated in \prettyref{fig:CompareDesign}, our SeedGNN with full design achieves the best performance among all variants, which shows the effectiveness of our design choices for the SeedGNN architecture. Further, among the variants, SeedGNN-w almost fails completely, which highlights the significant importance of using the percolation idea in SeedGNN for seeded graph matching. SeedGNNx does performs poorly, which demonstrates that it is also important to extract witness-like information at a larger number of hops instead of only 1-hop.  We can observe that SeedGNN and SeedGNN-h both outperform SeedGNN-p and the improvement of SeedGNN is significantly bigger. This result verifies that it is not enough to only use the soft-correspondence (as in SeedGNN-p), and we need to combine both the matching result $R_l$ of the Hungarian algorithm and the similarity $Y_l$ as in \prettyref{eq:mask} to achieve the best performance. 

\subsection{Study of the  Necessary  Training Samples for Generalization}
\begin{figure}[ht]
\centering
\begin{subfigure}{0.33\textwidth}
\centering
\includegraphics[scale=0.35]{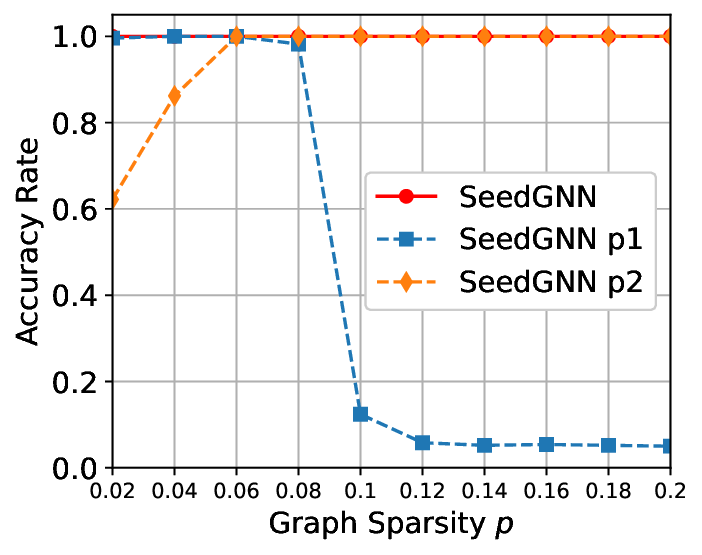}
\caption{$s=0.8,\ \theta = 0.05$.}
\label{fig:Comparesparsity}
\end{subfigure}\hfill
\begin{subfigure}{0.33\textwidth}
\centering
\includegraphics[scale=0.35]{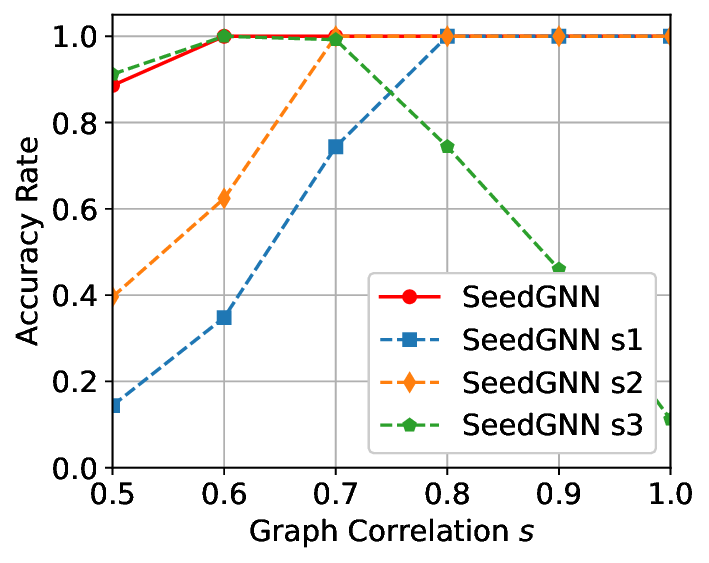}
\caption{$p=0.08,\ \theta = 0.05$.}
\label{fig:Comparecorrelation}
\end{subfigure}\hfill
\begin{subfigure}{0.33\textwidth}
\centering
\includegraphics[scale=0.35]{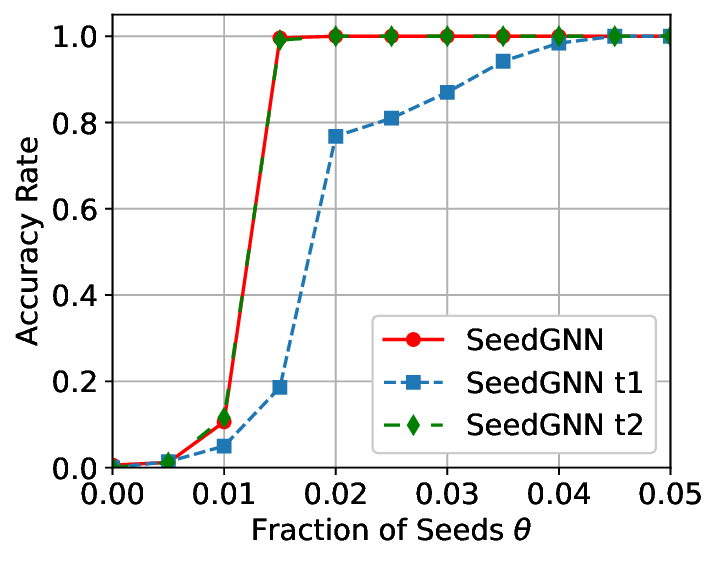}
\caption{$p=0.04,\ s=0.8$.}
\label{fig:Comparetheta}
\end{subfigure}
\caption{Performance comparison of SeedGNN trained with different training sets. Fix $n=500$.}
\label{fig:Comparenecessary}
\end{figure}

Intuitively, in order to help our SeedGNN successfully learn useful knowledge that can be applied to never-seen graphs, the training set needs to contain graph pairs with different varieties, e.g., graph sparsity, graph correlation, and the size of seed set. However, a larger training set also increases the training time.  To show which sets of graph pairs are necessary, we compare SeedGNN trained with different training sets, whose parameters are shown in \prettyref{tab:trainingset}. We use $\mathcal{T}$ to denote the training set that only includes the  \ER graphs of the training set in \prettyref{sec:exp-set}. First, to show the necessity of training graph pairs with a wide range of sparsity, we train  SeedGNN  with $\mathcal{T}$, $\mathcal{T}_{p1}$ and $\mathcal{T}_{p2}$, and compare the performance of the trained models while increasing $p$ from $0.02$ to $0.2$ and fixing $n=500$, $s=0.8$ and $\theta = 0.05$. \prettyref{fig:Comparesparsity} shows that, if SeedGNN is only trained with $p=0.1$, it performs well on sparse graphs but poorly on dense graphs. In contrast, if SeedGNN is only trained with $p=0.5$, it performs well on dense graphs but poorly on sparse graphs. Thus, we should include both $p=0.1$ and $p=0.5$ in the training set to achieve good performance. Second, to show the necessity of training graph pairs with different correlations, we compare the performance of  SeedGNN trained with $\mathcal{T}$, $\mathcal{T}_{s1}$, $\mathcal{T}_{s2}$ and $\mathcal{T}_{s3}$, and compare these models while increasing $s$ from $0.5$ to $1$ and fixing $n=500$, $p=0.08$ and $\theta = 0.05$. \prettyref{fig:Comparecorrelation} shows that, if SeedGNN is only trained with $s=0.6$, it performs well on  moderately correlated graphs but poorly on highly correlated graphs. In contrast, if SeedGNN is only trained with $s=0.8$ or $s=1$, it performs well on highly correlated graphs but poorly on moderately correlated  graphs. Thus, we should include different correlations in the training set to achieve good performance. Third,  we compare the performance of  SeedGNN trained with $\mathcal{T}$, $\mathcal{T}_{t1}$ and $\mathcal{T}_{t2}$, and compare these models while increasing $\theta$ from $0$ to $0.05$ and fixing $n=500$, $p=0.04$ and $s = 0.8$. \prettyref{fig:Comparetheta} shows that, if SeedGNN is only trained with $\theta=0.1$ and $\theta\in\{0.1,0.3\}$, it performs exactly the same. If SeedGNN is  only trained with $\theta = 0.3$, it performs worse than the former two. Thus, we only need to include graph pairs with a relatively small seed set in the training set.
\begin{table}[ht]
\begin{center}
\small
\caption{Different Training Sets}
\def \temptablewidth {0.9\textwidth}
\begin{tabular}{|c|c|c|c|c|}
\hline
Training Sets & $p$ & $s$ & $\theta$  \\
\hline
$\mathcal{T}_{p1}$ & $\{0.1\}$ & $\{0.6,0.8,1\}$ & $\{0.05,0.1\}$ \\
\hline$\mathcal{T}_{p2}$ & $\{0.5\}$ & $\{0.6,0.8,1\}$ & $\{0.05,0.1\}$ \\
\hline$\mathcal{T}_{s1}$ & $\{0.1, 0.5\}$ & $\{1\}$ & $\{0.05,0.1\}$ \\
\hline$\mathcal{T}_{s2}$ &  $\{0.1, 0.5\}$ & $\{0.8\}$ & $\{0.05,0.1\}$ \\
\hline$\mathcal{T}_{s3}$ &  $\{0.1, 0.5\}$ & $\{0.6\}$ & $\{0.05,0.1\}$ \\
\hline$\mathcal{T}_{t1}$ &  $\{0.1, 0.5\}$ & $\{0.6,0.8,1\}$ & $\{0.3\}$ \\
\hline$\mathcal{T}_{t2}$ & $\{0.1, 0.5\}$ & $\{0.6,0.8,1\}$ & $\{0.1,0.3\}$ \\
\hline
\end{tabular}
\label{tab:trainingset}
\end{center}
\end{table}

\subsection{Layer-wise Study of SeedGNN}\label{sec:layer-wise}

Recall from \prettyref{sec:percolation} that our design on the feature combination potentially enables SeedGNN to utilize various types of witness-like information adaptively, based on the confidence levels of new seeds $z_l$. In this section, we verify this capability through numerical results. 
To directly visualize $z_l$ in the matching process, we present the similarity matrix $Y_l$ of each layer of SeedGNN and compare it with the witness matrix of the iterative 1-hop and 2-hop algorithms at each iteration. We assume that the true mapping $\pi$ is the identity permutation, \ie, $\pi(i)=i$.

\begin{figure}[ht]
\centering
\begin{subfigure}{0.16\textwidth}
\centering
\includegraphics[scale=0.24]{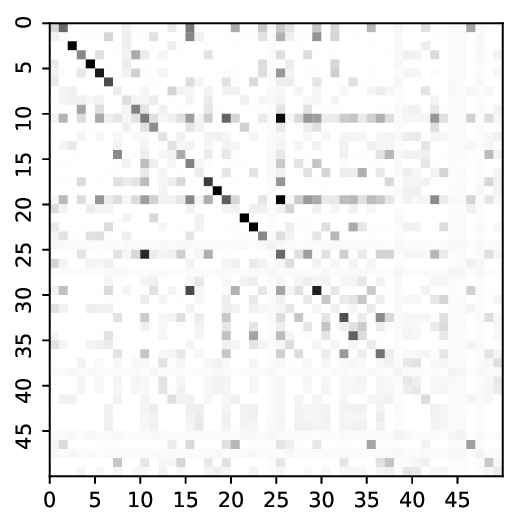}
\caption{Layer 1.}
\label{fig:ERmatrixdensegnn1}
\end{subfigure}
\begin{subfigure}{0.16\textwidth}
\centering
\includegraphics[scale=0.24]{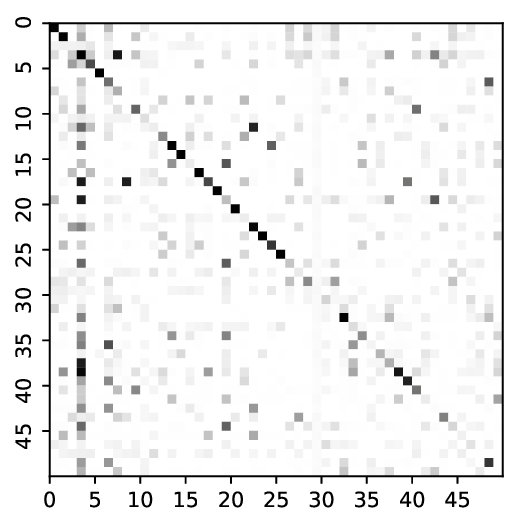}
\caption{Layer 2.}
\label{fig:ERmatrixdensegnn2}
\end{subfigure}
\begin{subfigure}{0.16\textwidth}
\centering
\includegraphics[scale=0.24]{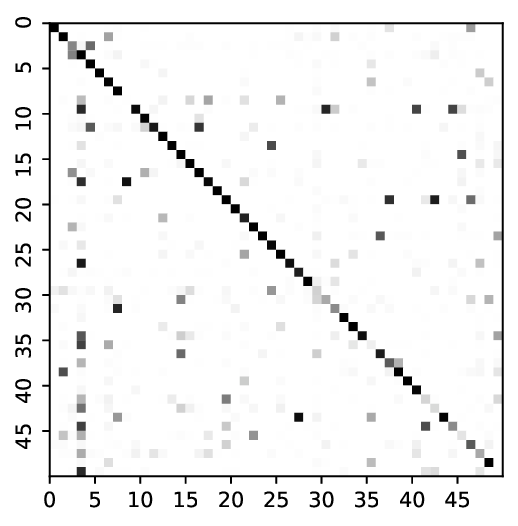}
\caption{Layer 3.}
\label{fig:ERmatrixdensegnn3}
\end{subfigure}
\begin{subfigure}{0.16\textwidth}
\centering
\includegraphics[scale=0.24]{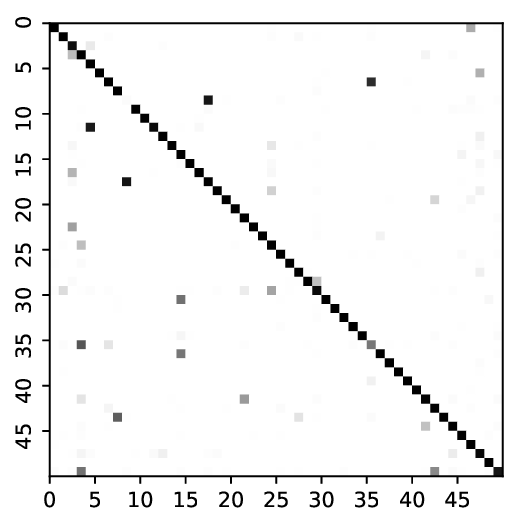}
\caption{Layer 4.}
\label{fig:ERmatrixdensegnn4}
\end{subfigure}
\begin{subfigure}{0.16\textwidth}
\centering
\includegraphics[scale=0.24]{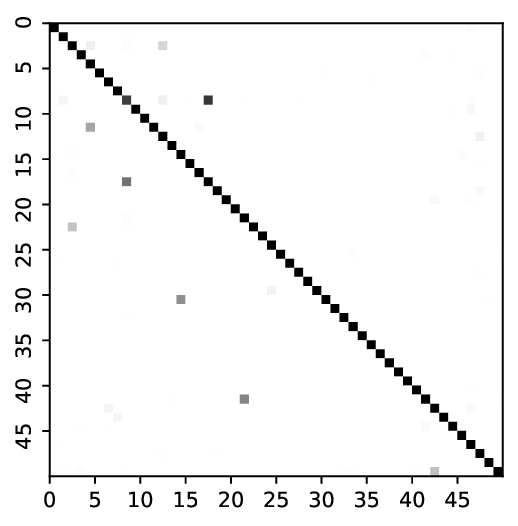}
\caption{Layer 5.}
\label{fig:ERmatrixdensegnn5}
\end{subfigure}
\begin{subfigure}{0.16\textwidth}
\centering
\includegraphics[scale=0.24]{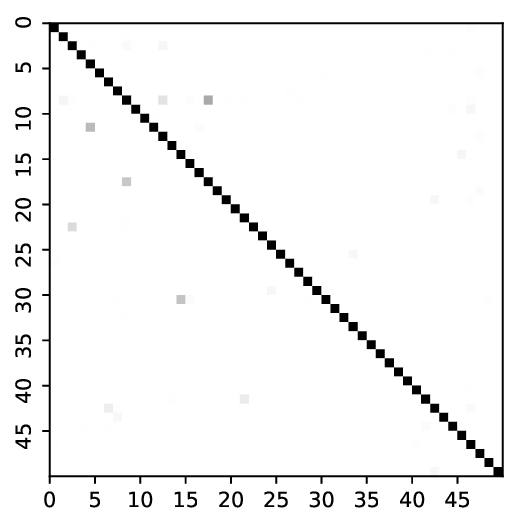}
\caption{Layer 6.}
\label{fig:ERmatrixdensegnn6}
\end{subfigure}
\begin{subfigure}{0.16\textwidth}
\centering
\includegraphics[scale=0.24]{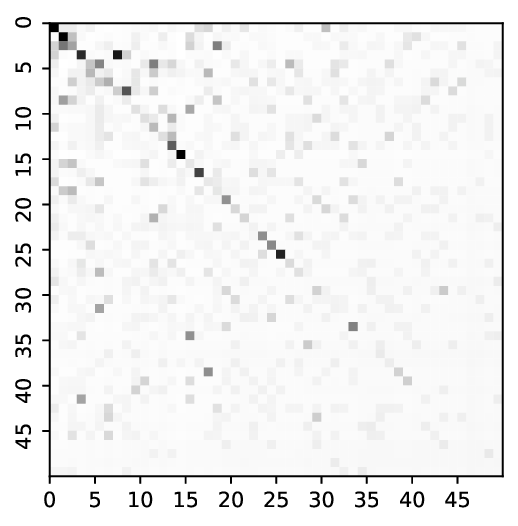}
\caption{Iteration 1.}
\label{fig:ERmatrix1}
\end{subfigure}
\begin{subfigure}{0.16\textwidth}
\centering
\includegraphics[scale=0.24]{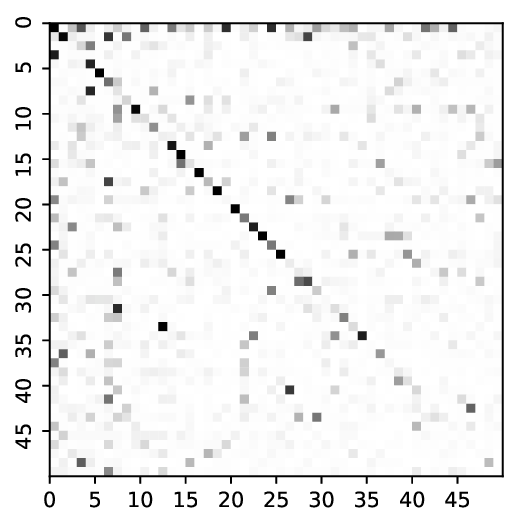}
\caption{Iteration 2.}
\label{fig:ERmatrix2}
\end{subfigure}
\begin{subfigure}{0.16\textwidth}
\centering
\includegraphics[scale=0.24]{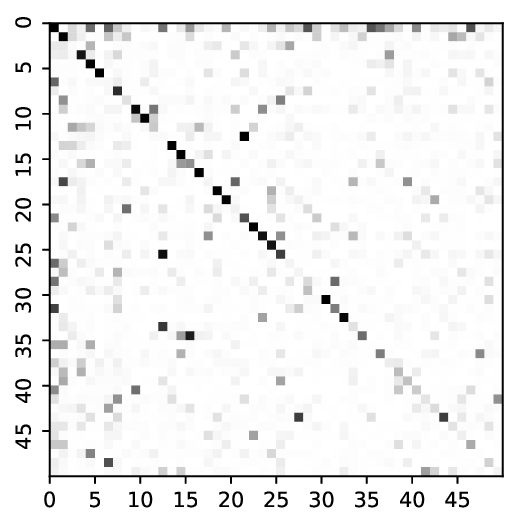}
\caption{Iteration 3.}
\label{fig:ERmatrix3}
\end{subfigure}
\begin{subfigure}{0.16\textwidth}
\centering
\includegraphics[scale=0.24]{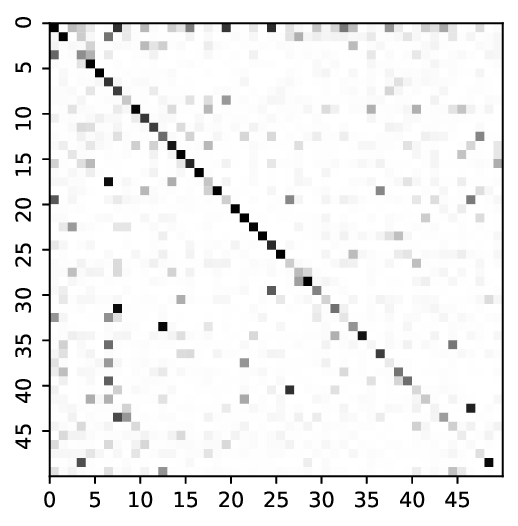}
\caption{Iteration 4.}
\label{fig:ERmatrix4}
\end{subfigure}
\begin{subfigure}{0.16\textwidth}
\centering
\includegraphics[scale=0.24]{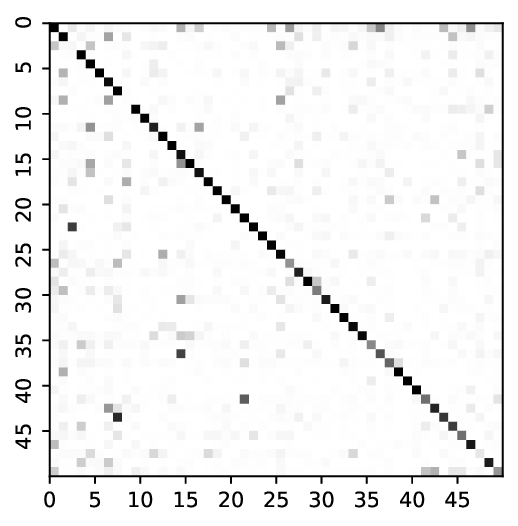}
\caption{Iteration 5.}
\label{fig:ERmatrix5}
\end{subfigure}
\begin{subfigure}{0.16\textwidth}
\centering
\includegraphics[scale=0.24]{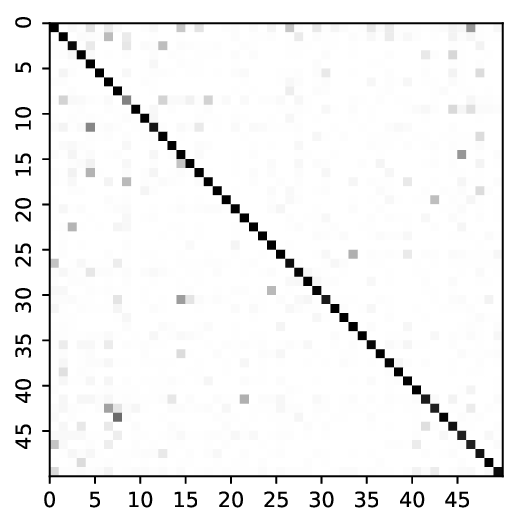}
\caption{Iteration 6.}
\label{fig:ERmatrix6}
\end{subfigure}
\begin{subfigure}{0.16\textwidth}
\centering
\includegraphics[scale=0.24]{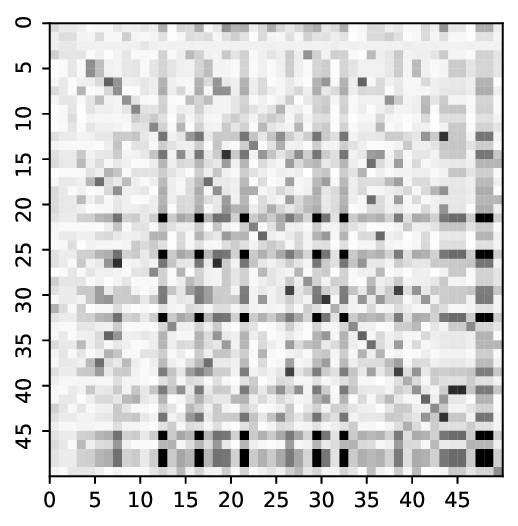}
\caption{Iteration 1.}
\label{fig:ERmatrix2hop1}
\end{subfigure}\hspace{2.83cm}
\begin{subfigure}{0.16\textwidth}
\centering
\includegraphics[scale=0.24]{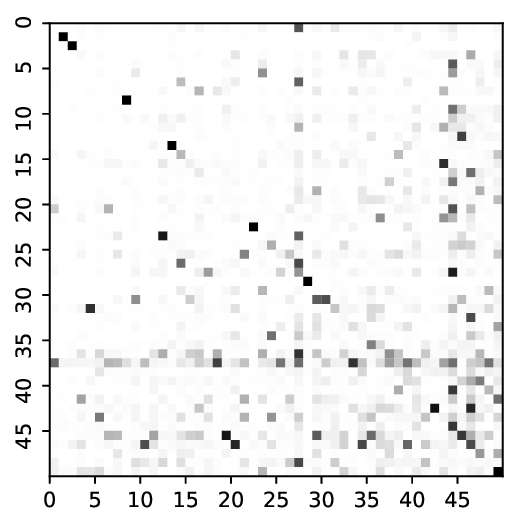}
\caption{Iteration 2.}
\label{fig:ERmatrix2hop2}
\end{subfigure}\hspace{2.83cm}
\begin{subfigure}{0.16\textwidth}
\centering
\includegraphics[scale=0.24]{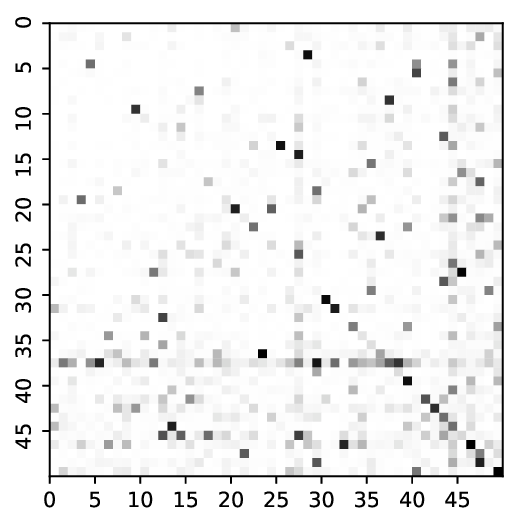}
\caption{Iteration 3.}
\label{fig:ERmatrix2hop3}
\end{subfigure}\hspace{-2.83cm}
\caption{ The similarity/witness matrices of the matching process on a fixed pair of dense correlated \ER graphs with $n=50$, $p=0.4$, $s=0.8$ and $\theta = 0.1$. Darker points correspond to higher similarity (in $Y_l$) or a larger number of witnesses. \prettyref{fig:ERmatrixdensegnn1} --- \prettyref{fig:ERmatrixdensegnn6} are the similarity matrix from each layer of SeedGNN. \prettyref{fig:ERmatrix1} --- \prettyref{fig:ERmatrix6} are the witness matrix from each iteration of the iterative 1-hop algorithm. \prettyref{fig:ERmatrix2hop1} --- \prettyref{fig:ERmatrix2hop3} are the witness matrix from each iteration of the iterative 2-hop algorithm.}
\label{fig:matrixdense}
\end{figure}

First, we study the matching process in \emph{dense} graphs. We fix a pair of correlated \ER graphs with $n=50$, $p=0.4$, $s=0.8$ and $\theta = 0.1$. Then, we index the nodes from 0 to 49 in the descending order of the node degree in the parent graph $\mathcal{G}_0$. In \prettyref{fig:matrixdense}, we show the similarity matrix $Y_l$ in each layer of our SeedGNN, and compare it with the witness matrix in each iteration using either the 1-hop or 2-hop algorithm. We can immediately see that the similarity matrices provided by SeedGNN are more similar to the witness matrices of the iterative 1-hop algorithm than that of the iterative 2-hop algorithm. Specifically, since the graphs are dense, the 1-hop witness from the initial seeds can already generate new seeds with high confidence levels (see \prettyref{fig:ERmatrixdensegnn1} and \ref{fig:ERmatrix1}, where there are many dark points on the diagonal (i.e., consistent with the underlying true mapping), while there are few dark points off the diagonal). The iterative 1-hop algorithm is known to use new 1-hop witnesses from these new seeds (see \prettyref{fig:ERmatrix2}) in the next iteration. In contrast, the 2-hop witnesses from the initial seeds are much noisier (see \prettyref{fig:ERmatrix2hop1}, where the darkness of the points on the diagonal cannot be differentiated from those off the diagonal). As we illustrated in \prettyref{fig:layerwise-feature}, these two types of witness-like information are both contained in the second layer of SeedGNN. 
By comparing \prettyref{fig:ERmatrixdensegnn2} with \prettyref{fig:ERmatrix2} and \prettyref{fig:ERmatrix2hop1}, we can observe that the second layer of SeedGNN produces a similarity matrix that is closer to the witness matrix of the 1-hop algorithm than that of the 2-hop algorithm. Thus, we infer that, for these dense graphs in which the new seeds are reliable, the SeedGNN relies more on witnesses computed from these new seeds. 


\begin{figure}[ht]
\centering
\begin{subfigure}{0.16\textwidth}
\centering
\includegraphics[scale=0.24]{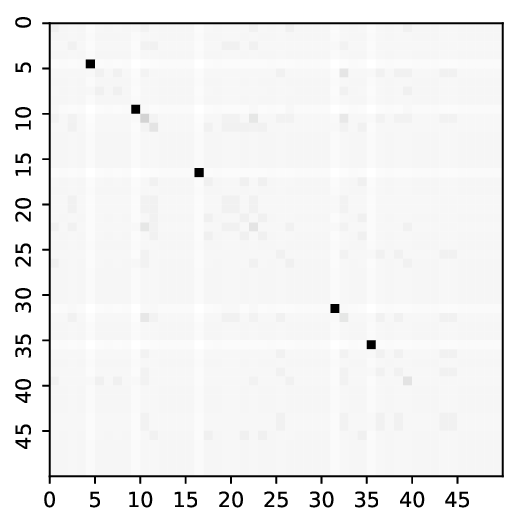}
\caption{Layer 1.}
\label{fig:ERmatrixsparsegnn1}
\end{subfigure}
\begin{subfigure}{0.16\textwidth}
\centering
\includegraphics[scale=0.24]{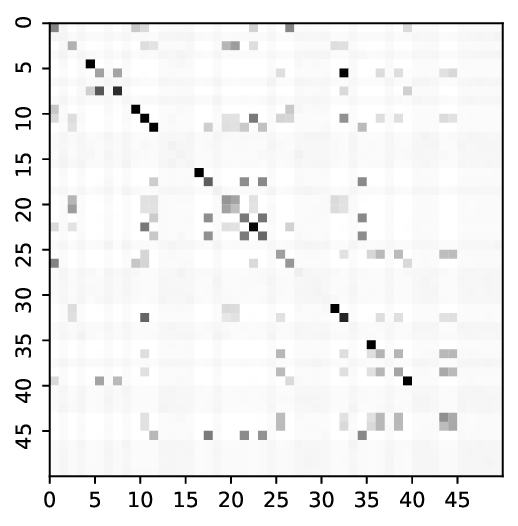}
\caption{Layer 2.}
\label{fig:ERmatrixsparsegnn2}
\end{subfigure}
\begin{subfigure}{0.16\textwidth}
\centering
\includegraphics[scale=0.24]{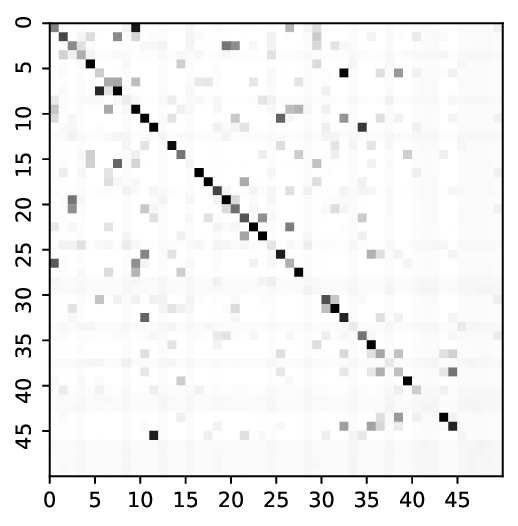}
\caption{Layer 3.}
\label{fig:ERmatrixsparsegnn3}
\end{subfigure}
\begin{subfigure}{0.16\textwidth}
\centering
\includegraphics[scale=0.24]{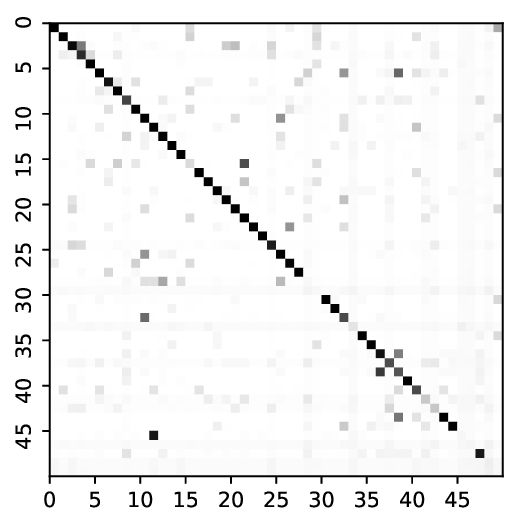}
\caption{Layer 4.}
\label{fig:ERmatrixsparsegnn4}
\end{subfigure}
\begin{subfigure}{0.16\textwidth}
\centering
\includegraphics[scale=0.24]{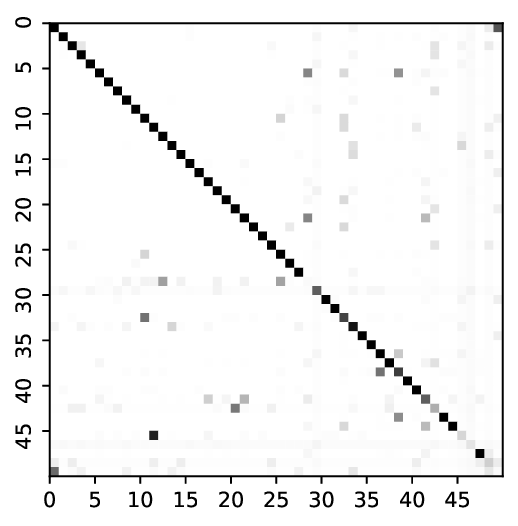}
\caption{Layer 5.}
\label{fig:ERmatrixsparsegnn5}
\end{subfigure}
\begin{subfigure}{0.16\textwidth}
\centering
\includegraphics[scale=0.24]{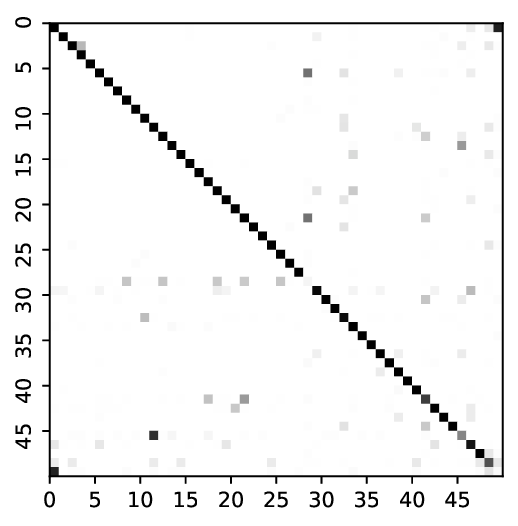}
\caption{Layer 6.}
\label{fig:ERmatrixsparsegnn6}
\end{subfigure}
\begin{subfigure}{0.16\textwidth}
\centering
\includegraphics[scale=0.24]{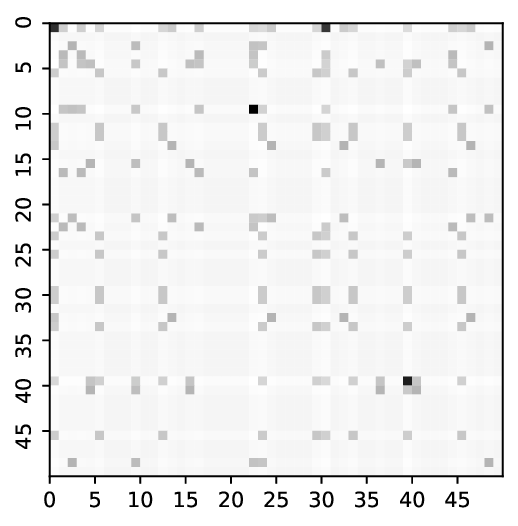}
\caption{Iteration 1.}
\label{fig:ERmatrixsparse1}
\end{subfigure}
\begin{subfigure}{0.16\textwidth}
\centering
\includegraphics[scale=0.24]{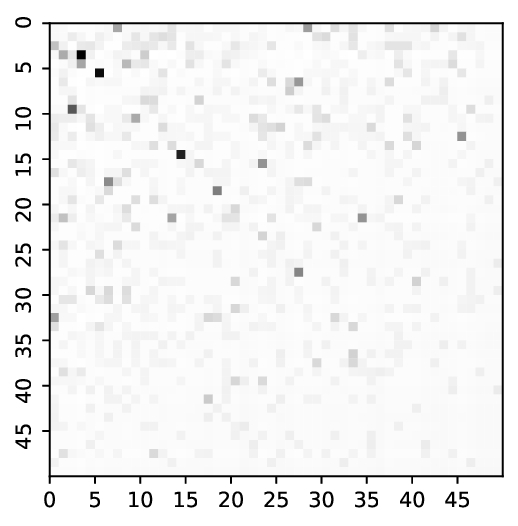}
\caption{Iteration 2.}
\label{fig:ERmatrixsparse2}
\end{subfigure}
\begin{subfigure}{0.16\textwidth}
\centering
\includegraphics[scale=0.24]{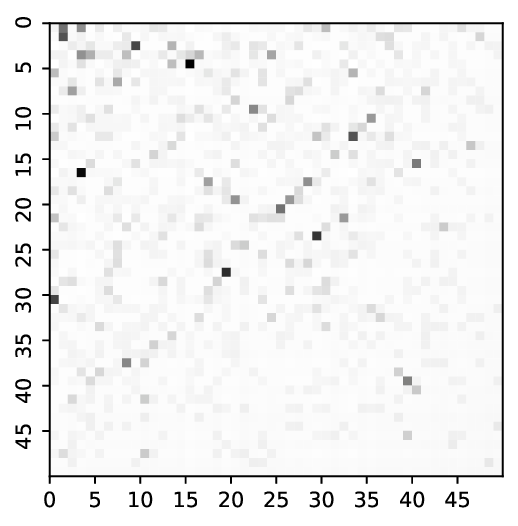}
\caption{Iteration 3.}
\label{fig:ERmatrixsparse3}
\end{subfigure}
\begin{subfigure}{0.16\textwidth}
\centering
\includegraphics[scale=0.24]{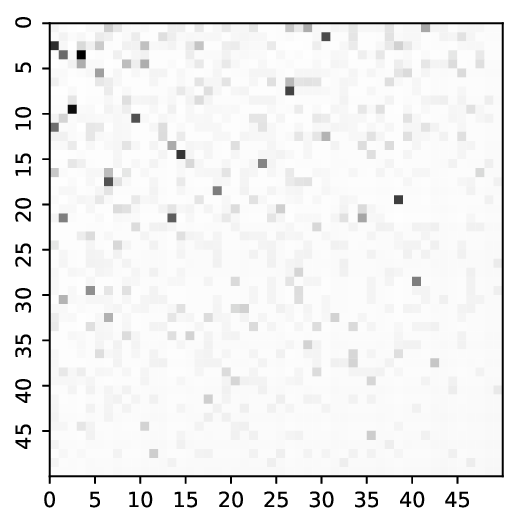}
\caption{Iteration 4.}
\label{fig:ERmatrixsparse4}
\end{subfigure}
\begin{subfigure}{0.16\textwidth}
\centering
\includegraphics[scale=0.24]{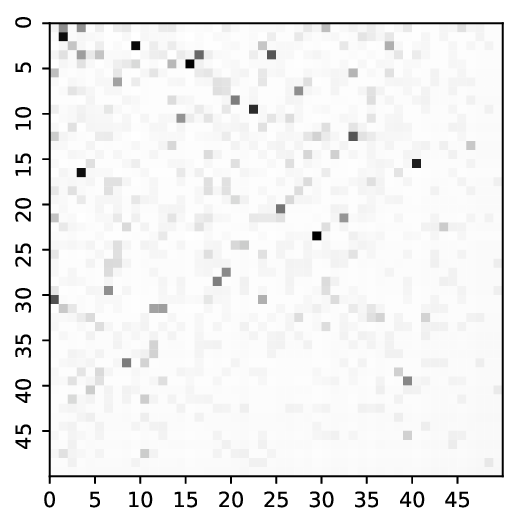}
\caption{Iteration 5.}
\label{fig:ERmatrixsparse5}
\end{subfigure}
\begin{subfigure}{0.16\textwidth}
\centering
\includegraphics[scale=0.24]{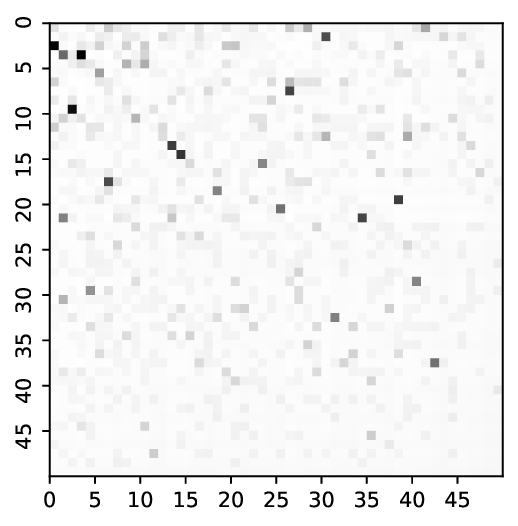}
\caption{Iteration 6.}
\label{fig:ERmatrixsparse6}
\end{subfigure}
\begin{subfigure}{0.16\textwidth}
\centering
\includegraphics[scale=0.24]{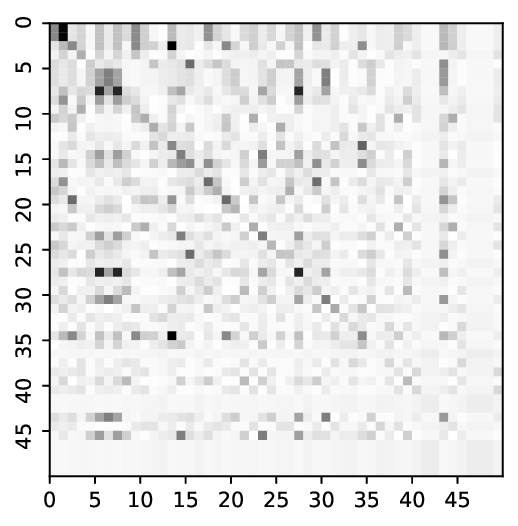}
\caption{Iteration 1.}
\label{fig:ERmatrixsparse2hop1}
\end{subfigure}\hspace{2.83cm}
\begin{subfigure}{0.16\textwidth}
\centering
\includegraphics[scale=0.24]{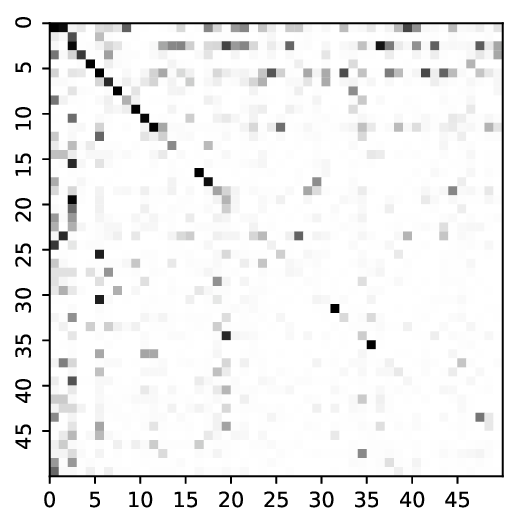}
\caption{Iteration 2.}
\label{fig:ERmatrixsparse2hop2}
\end{subfigure}\hspace{2.83cm}
\begin{subfigure}{0.16\textwidth}
\centering
\includegraphics[scale=0.24]{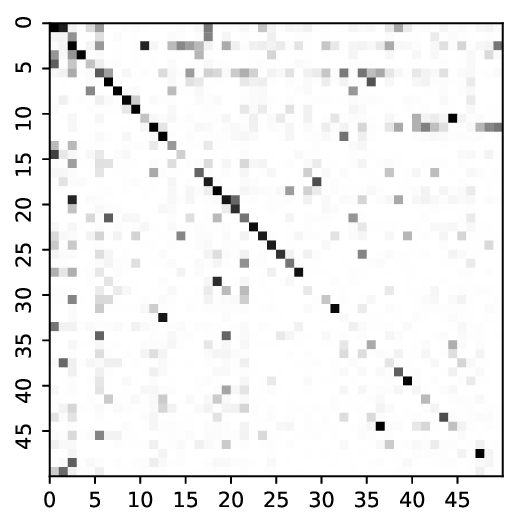}
\caption{Iteration 3.}
\label{fig:ERmatrixsparse2hop3}
\end{subfigure}\hspace{-2.83cm}
\caption{The similarity/witness matrices of the matching process on a fixed pair of sparse correlated \ER graphs with $n=50$, $p=0.1$, $s=0.8$ and $\theta = 0.1$. Darker points correspond to higher similarity (in $Y_l$) or a larger number of witnesses. \prettyref{fig:ERmatrixsparsegnn1} --- \prettyref{fig:ERmatrixsparsegnn6} are the similarity matrix from each layer of SeedGNN. \prettyref{fig:ERmatrixsparse1} --- \prettyref{fig:ERmatrixsparse6} are the witness matrix from each iteration of the iterative 1-hop algorithm. \prettyref{fig:ERmatrixsparse2hop1} --- \prettyref{fig:ERmatrixsparse2hop3} are the witness matrix from each iteration of the iterative 2-hop algorithm.}
\label{fig:matrixsparse}
\end{figure}

Then, we study the matching process in \emph{sparse} graphs. We fix a pair of correlated \ER graphs with $n=50$, $p=0.1$, $s=0.8$ and $\theta = 0.1$. Then, we also index the nodes from 0 to 49 in the descending order of the node degree in the parent graph $\mathcal{G}_0$. In \prettyref{fig:matrixsparse}, we show the similarity matrix $Y_l$ in each layer of our SeedGNN, and compare it with the witness matrix in each iteration using either the 1-hop or 2-hop algorithm. 
In contrast to \prettyref{fig:matrixdense}, in this case, we observe that the similarity matrices provided by SeedGNN are more similar to the witness matrices of the iterative 2-hop algorithm than those of the iterative 1-hop algorithm. Specifically, since the graphs are sparse, there are very few 1-hop witnesses even for true pairs. Thus, the 1-hop algorithm almost fails completely (see \prettyref{fig:ERmatrixsparse1} --- \prettyref{fig:ERmatrixsparse6}). On the contrary, the 2-hop witnesses from the initial seeds are much more reliable (see \prettyref{fig:ERmatrixsparse2hop1}). As a result, the iterative 2-hop algorithm produces much better results (see \prettyref{fig:ERmatrixsparse2hop1} --- \prettyref{fig:ERmatrixsparse2hop3}). 
By comparing \prettyref{fig:ERmatrixsparsegnn2} with \prettyref{fig:ERmatrixsparse2} and \prettyref{fig:ERmatrixsparse2hop1}, we can observe that the second layer of SeedGNN produces a similarity matrix that is closer to the witness matrix of the 2-hop algorithm than that of the 1-hop algorithm. Thus, we can infer that, for these sparse graphs in which the confidence levels of new seeds are low, SeedGNN utilizes 2-hop witness-like information from the initial seeds, and avoids using 1-hop witnesses based on these new seeds. 

In summary, from these two case studies, we conclude that our SeedGNN might be able to choose the appropriate features for different types of graphs according to the confidence level of new seeds.
Further, we observe that the matching accuracy of SeedGNN is even higher than that of the 1-hop and 2-hop algorithms, the latter two of which have been theoretically proven to work well for dense graphs and sparse graphs, respectively \cite{mossel2019seeded}.  Thus, this result suggests that SeedGNN may extract more valuable features, or learn more effective ways to synthesize witness-like information, than the theoretical algorithms.



\section{Limitations}\label{app:limitations}
The limitations of our proposed SeedGNN are three-fold. 1) We only consider using topological structure in our SeedGNN. Although the non-topological features are sometimes hard to obtain or inaccurate due to various constraints in practice, when they are available, we may consider combining them with topological features. 2) We only adopt the insights from theoretical algorithms for seeded graph matching. The seedless graph matching algorithms may yield some additional useful insights on effectively using the topological information. 3) Compared to non-learning methods, SeedGNN requires additional training process and higher computational complexity.  

There are many interesting future directions to overcome the limitations, such as simultaneously using topological and non-topological features, and extending our key ideas into seedless graph matching.

\section{Broader Impact}\label{app:impact}

Our work has a positive impact on many practical applications, such as network privacy, computational biology, computer vision, and natural language processing. For example, with the increasing number and scale of social networks, mapping users across online social networks attracts much attention from both academia and industry. Our work can further help information analysis, such as user behavior prediction \cite{10.5555/3015812.3015815}, identity verification and cross-domain recommendation \cite{lu2016item,li2014matching}. Another important application of graph matching is protein interaction network alignment \cite{singh2008global,kazemi2016proper,kriege2019chemical}. The alignment of protein-protein interaction (PPI) networks enables us to uncover the relationships between different species, which leads to a deeper understanding of biological systems. In computer vision, our work can be applied in finding similar images \cite{conte2004thirty,schellewald2005probabilistic,vento2013graph} and matching 3D deformable shapes~\cite{kim2011blended,lahner2016shrec,vestner2017efficient,vestner2017product}. In natural language processing, our work can be used in question answering, machine translation, and information retrieval \cite{haghighi2005robust}.

Participants joining different social platforms may have privacy or anonymity considerations.  Our work may have some adverse impact on user privacy protection. However, we believe that our proposed design choices may also be useful for guiding the design of privacy protection schemes.

\end{document}

%% file: math_commands.tex
\usepackage{amsmath,amsfonts,bm}

\def\eqref#1{equation~\ref{#1}}

\def\1{\bm{1}}

\DeclareMathAlphabet{\mathsfit}{\encodingdefault}{\sfdefault}{m}{sl}
\SetMathAlphabet{\mathsfit}{bold}{\encodingdefault}{\sfdefault}{bx}{n}

